\documentclass[conference]{IEEEtran}
\pdfoutput=1 
\newtheorem{theorem}{Theorem}[section]

\newtheorem{lem}[theorem]{Lemma}
\newtheorem{prop}[theorem]{Proposition}
\newtheorem{assumption}[theorem]{Assumption}
\newtheorem{problem}{Problem}
\newtheorem{definition}[theorem]{Definition}
\newtheorem{rem}[theorem]{Remark}
\newtheorem{ex}[theorem]{Example}

\usepackage{graphicx} 
\usepackage{epsfig} 
\usepackage{amsmath}
\usepackage{amssymb}
\usepackage{epstopdf}
\usepackage{cite}
\usepackage[noend,ruled,linesnumbered]{algorithm2e}
\SetKwComment{Comment}{$\triangleright$\ } {}
\usepackage{multirow}
\usepackage{rotating}
\usepackage{subfigure} 
\usepackage{color} 
\usepackage{mysymbol}
\usepackage[dvipsnames]{xcolor}
\usepackage[hyphens]{url}

\usepackage[breaklinks=true, colorlinks, bookmarks=true, citecolor=Black, urlcolor=Violet,linkcolor=Black]{hyperref}

\usepackage{comment}
\usepackage[colorinlistoftodos,prependcaption,textwidth=1.5cm,textsize=tiny]{todonotes}

\IEEEoverridecommandlockouts                              

\begin{document}
\title{\LARGE \bf Reactive Temporal Logic Planning for Multiple Robots \\
in Unknown Occupancy Grid Maps}
\author{Yiannis Kantaros, Matthew Malencia, and George J. Pappas\thanks{The authors are with with the GRASP Laboratory, University of Pennsylvania, Philadelphia, PA, 19104, USA. $\left\{\text{kantaros, malencia, pappasg}\right\}$@seas.upenn.edu.  
This work was supported by DARPA Assured Autonomy and AFOSR grant FA9550-19-1-0265. 
}}
\maketitle 

\begin{abstract}
This paper proposes a new reactive temporal logic planning algorithm for multiple robots that operate in environments with unknown geometry modeled using occupancy grid maps. The robots are equipped with individual sensors that allow them to continuously learn a grid map of the unknown environment using existing Simultaneous Localization and Mapping (SLAM) methods. The goal of the robots is to accomplish complex collaborative tasks, captured by global Linear Temporal Logic (LTL) formulas. The majority of existing LTL planning approaches rely on discrete abstractions of the robot dynamics operating in {\em known} environments and,  as a result, they cannot be applied to the more realistic scenarios where the environment is initially unknown. In this paper, we address this novel challenge by proposing the first \textit{reactive}, \textit{abstraction-free}, and \textit{distributed} LTL planning algorithm that can be applied for complex mission planning of multiple robots operating in unknown environments. The proposed algorithm is {\em reactive} i.e., planning is adapting to the updated environmental map and \textit{abstraction-free} as it does not rely on designing abstractions of the robot dynamics. Also, our algorithm is {\em distributed} in the sense that the global LTL task is decomposed into single-agent reachability problems constructed online based on the continuously learned map. The proposed algorithm is complete under mild assumptions on the structure of the environment and the sensor models. We provide extensive numerical simulations and hardware experiments that illustrate the theoretical analysis and show that the proposed algorithm can address complex planning tasks for large-scale multi-robot systems in unknown environments.
\end{abstract}

%

   
\section{Introduction} \label{sec:Intro}

Temporal logic motion planning has emerged as one of the main approaches for specifying a richer class of robot tasks than the classical point-to-point navigation and can capture temporal and Boolean requirements, such as sequencing, surveillance or coverage
\cite{fainekos2005hybrid,kress2007s,smith2011optimal,tumova2016multi,chen2012formal,ulusoy2013optimality,ulusoy2014optimal,shoukry2017linear,shoukry2018smc,vasile2013sampling,kantaros2017Csampling,kantaros2018text}. Common in these works is that they typically assume robots with \textit{known} dynamics operating in \textit{known} environments that are modeled using discrete abstractions, e.g., transition systems \cite{baier2008principles}. 
As a result, these methods cannot be applied to scenarios where the environment is initially unknown and, therefore, online re-planning may be required as environmental maps are constructed;  resulting in limited applicability. To mitigate this issue, learning-based approaches have also been proposed that consider robots with \textit{unknown} dynamics operating in \textit{unknown} environments \cite{dorsa2014learning,li2017reinforcement,jones,gao2019reduced,hasanbeig2019reinforcement}. These approaches learn policies that directly map on-board sensor readings to control commands in a trial-and-error fashion. However, learning-based approaches, in addition to being data inefficient, are specialized to the  environment they were trained on and do not generalize well to previously unseen environments. 



This paper addresses a motion planning problem for multiple robots with \textit{known} dynamics that need to accomplish collaborative complex tasks in \textit{unknown} environments. We consider robots that are tasked with accomplishing collaborative tasks captured by global Linear Temporal Logic (LTL) formulas in complex environments with unknown geometric structure. To address this problem, we propose the first reactive, abstraction-free, and distributed temporal logic planning algorithm for complex mission planning in unknown environments. The robots are equipped with sensors that allow them to continuously learn an individual or shared occupancy grid map of the environment, using existing Simultaneous Localization and Mapping (SLAM) methods \cite{elfes1989using,thrun2001learning,thrun2005probabilistic,grisetti2007improved}, while reacting and adapting mission planning to the updated map.
Furthermore, the proposed algorithm is abstraction-free i.e., it does not rely on any discrete abstraction of the robot dynamics (e.g., transition systems) and distributed in the sense that the temporal logic task is decomposed into single-agent point-to-point navigation tasks in the presence of unknown obstacles. These local tasks are formulated collaboratively and online based on the continuously learned map while they can be solved using existing approaches \cite{karaman2011sampling,guzzi2013human,liu2016high,otte2016rrtx,bansal2019combining,vasilopoulos2018sensor,ryll2019efficient,tordesillas2020faster}.
The decomposition of the global LTL task into local reachability tasks allows for efficient re-planning, as the robots can make decisions locally to accomplish the assigned LTL task while communicating intermittently with each other when new reachability tasks need to be designed. 
Theoretically, we show there exists a fragment of LTL formulas for which decomposition into local reachability tasks is always possible. Furthermore, we show that the paths designed to address these reachability sub-tasks can be executed asynchronously across the robots by introducing waiting actions. We also prove that under mild assumptions on the coarse structure of the environment and the sensor model, the proposed planning algorithm is complete. Finally, we validate the proposed safe planning algorithm in both  simulations to demonstrate its scalability and in hardware experiments to test its performance when assumptions on sensor models do not hold. 

\subsection{Related Research}

Reactive temporal logic planning algorithms that can account for map uncertainty in terms of incomplete environment models have been developed  in \cite{guo2013revising,guo2015multi,maly2013iterative,lahijanian2016iterative,livingston2012backtracking,livingston2013patching,kress2009temporal,alonso2018reactive}. Particularly, \cite{guo2013revising,guo2015multi} model the environment as transition systems which are partially known. Then discrete controllers are designed by applying graph search methods on a product automaton. As the environment, i.e., the transition system, is updated, the product automaton is locally updated as well, and new paths are re-designed by applying graph search approaches on the revised automaton. 
A conceptually similar approach is proposed in \cite{maly2013iterative,lahijanian2016iterative} as well. The works in \cite{livingston2012backtracking,livingston2013patching} propose methods to locally patch paths as transition systems, modeling the environment, change so that GR(1) (General Reactivity of Rank 1) specifications are satisfied. Reactive to LTL specifications planning algorithms are proposed in \cite{kress2009temporal,alonso2018reactive}, as well. Specifically, in \cite{kress2009temporal,alonso2018reactive} the robot reacts to the environment while the task specification captures this reactivity. Correctness of these algorithms is guaranteed if the robot operates in an environment that satisfies the assumptions that were explicitly modeled in the task specification. 
Common in all these works is that, unlike our approach, they rely on discrete abstractions of the robot dynamics  \cite{belta2005discrete,pola2008approximately}. Thus, correctness of the generated motion plans is guaranteed with respect to the discrete abstractions resulting in a gap between  the generated discrete motion plans and their physical low-level implementation.
Moreover, the reactive algorithms proposed in \cite{guo2013revising,guo2015multi,maly2013iterative,lahijanian2016iterative,livingston2012backtracking,livingston2013patching,kress2009temporal,alonso2018reactive} are centralized as they rely on constructing a product automaton among all robots to synthesize motion plans. As a result, their computational cost for online re-planning increases as the number of robots, the complexity of the task, or the size of the environment increase.
To the contrary, our algorithm is highly scalable due to its distributed nature and the fact that it avoids altogether the construction of a product automaton. 
Finally, note that decomposition of fragments of LTL formulas into reachability tasks has been investigated in \cite{fainekos2006translating,bisoffi2018hybrid,jagtap2019formal}, as well, assuming robots that operate in known environments. 
Common in these works is that temporal logic tasks are decomposed into \textit{global/multi-agent} reachability tasks. As a result, these methods face scaling challenges as the number of robots increases, as well, while they require synchronous motion among all agents for all time. Our proposed method mitigates these issues by decomposing global temporal logic tasks  into \textit{local single-agent} reachability tasks. 

%

\subsection{Contribution}
The contribution of this paper can be summarized as follows.
\textit{First}, we propose the \textit{first} formal bridge between temporal logic planning and SLAM methods. In fact, we propose the first algorithm for temporal logic planning in continuously learned occupancy grid maps which is also reactive, distributed, and abstraction-free in terms of the robot dynamics. \textit{Second}, we present a new approach to decompose global LTL tasks into single-agent reachability tasks which significantly decreases the computational cost of re-planning due to the unknown environmental structure. \textit{Third}, we show that the proposed algorithm is complete under certain assumptions on the coarse structure of the environment and the sensor model. \textit{Fourth}, we provide extensive comparative simulation studies and hardware experiments that corroborate the efficacy and scalability of the proposed algorithm.


\section{Problem Formulation} \label{sec:PF}

\begin{figure}[t]
  \centering
  \includegraphics[width=0.8\linewidth]{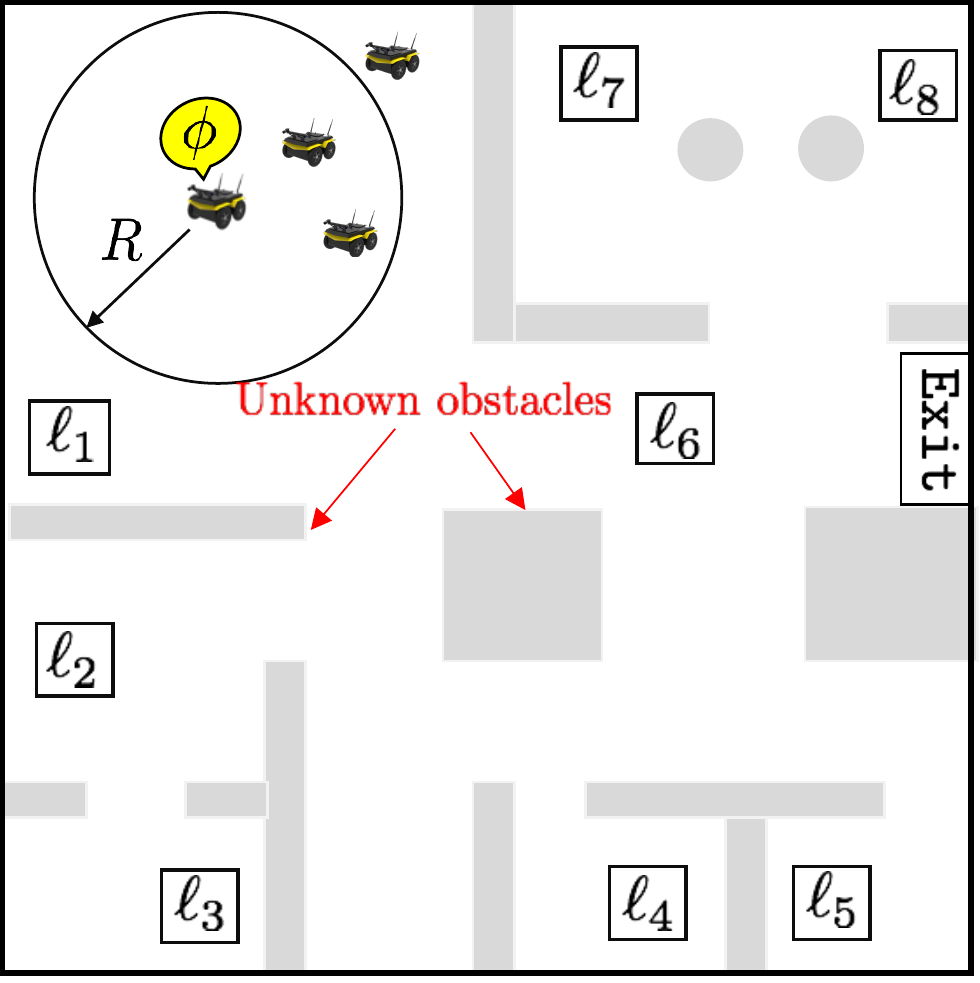}
  \caption{Illustration of the problem formulation. A team of robots with known dynamics, equipped with range-limited sensors, are responsible for accomplishing a collaborative mission, captured by a Linear Temporal Logic formula $\phi$, in an unknown environment. Specifically, the environment consists of known regions of interest $\ell_i$ and fully unknown obstacles/walls illustrated by gray polygons.
  }
  \label{fig:PF}
\end{figure}
\subsection{Multi-Robot Systems in Unknown Environments}
Consider $N$ mobile robots that reside in an environment $\Omega\subset \reals^d$ with non-overlapping and known regions of interest, denoted  by $\ell_i\subset \Omega$, and \textit{static} but \textit{unknown} obstacles $O\subset\Omega$, where $d$ is the dimension of the workspace; see also Figure \ref{fig:PF}. The robots are governed by the following dynamics: $\bbp_{j}(t+1)=\bbf_j(\bbp_{j}(t),\bbu_{j}(t))$,
%
for all $j\in\ccalN:=\{1,\dots,N\}$, where $\bbp_{j}(t) \in\mathbb{R}^n$ and $\bbu_{j}(t)$ stand for the state (e.g., position and orientation) and the control input of robot $j$ at time $t\geq 0$, respectively. 
Hereafter, we compactly denote the dynamics of all robots as 
\begin{equation}\label{eq:rdynamics}
\bbp(t+1)=\bbf(\bbp(t),\bbu(t)),
\end{equation}
where $\bbp(t)\in \mathbb{R}^{n\times N}$ and $\bbu(t)\in\ccalU:=\ccalU_1\times\dots\times\ccalU_N$, for all $t\geq 0$.

\subsection{Sensors \& Occupancy Grid Maps}
Moreover, we assume that the robots are equipped with  sensors (e.g., LiDAR or camera) to collect noisy range measurements $r$ that are used to detect obstacles. Typically, such sensors are modeled by a probability density function $p(r|z)$, where $z$ is the actual distance to the object being detected \cite{elfes1989using}.
%
%
%
Using noisy range sensor measurements, the robots collectively build a shared (binary) occupancy grid map using existing SLAM algorithms \cite{elfes1989using,thrun2001learning,thrun2005probabilistic,grisetti2007improved}. Specifically, the occupancy grid map is composed of $K$ grid cells denoted by $c_i\subset\Omega$. 
%
Hereafter,  we define the occupancy grid map as  $M(t)=\{m_1(t),m_2(t),\dots,m_K(t)\}$, where (i) $m_i(t)=1$ if the cell $c_i$ is believed to be occupied by an obstacle based on the collected measurements, and (ii) $m_i(t)=0$ if the cell $i$ is believed to be obstacle-free or unexplored.\footnote{In other words, the unexplored part of the environment is considered obstacle free.} Hereafter, for simplicity of notation, we assume that all robots build a shared map $M(t)$. This can be relaxed by allowing each robot to build its own local map $M_j(t)$, based on its individual sensor measurements.

\subsection{Mission \& Safety Specification}
The goal of the robots is to accomplish a complex collaborative task captured by a global Linear Temporal Logic (LTL) specification $\phi$. The assigned task $\phi$ is defined over a set of atomic propositions $\mathcal{AP}=\cup_{\forall j}\mathcal{AP}_j$, where $\mathcal{AP}_j=\cup_i\{\pi_{j}^{\ell_i},\pi_j^{O}\}$, and $\pi_{j}^{\ell_i}$ and $\pi_{j}^{O}$ are atomic predicates that are true when robot $j$ is within region $\ell_i$ and hits an obstacle, respectively.
In what follows, we briefly present the syntax and semantics of LTL. A detailed overview of this theory can be found in \cite{baier2008principles}. The basic ingredients of LTL are a set of atomic propositions $\mathcal{AP}$, the boolean operators, i.e., conjunction $\wedge$, and negation $\neg$, and two temporal operators, next $\bigcirc$ and until $\mathcal{U}$. LTL formulas over a set $\mathcal{AP}$ can be constructed based on the following grammar: $\phi::=\text{true}~|~\pi~|~\phi_1\wedge\phi_2~|~\neg\phi~|~\bigcirc\phi~|~\phi_1~\mathcal{U}~\phi_2$, where $\pi\in\mathcal{AP}$. For brevity we abstain from presenting the derivations of other Boolean and temporal operators, e.g., \textit{always} $\square$, \textit{eventually} $\lozenge$, \textit{implication} $\Rightarrow$, which can be found in \cite{baier2008principles}. 
Hereafter, we exclude the `next' operator $\bigcirc$ from the syntax, since it is not meaningful for practical robotics applications; this is common in relevant works, see, e.g., \cite{kloetzer2008fully} and the references therein. LTL formulas are satisfied by discrete plans $\tau$ that are
infinite sequences of locations of $N$ robots in $\Omega$. i.e., $\tau=\bbp(0),\bbp(1),\dots,\bbp(t),\dots$ \cite{baier2008principles}. 

An example of an LTL specification is: $\phi=\Diamond(\pi_1^{\ell_1}\wedge (\pi_1^{\ell_2}\wedge (\Diamond \pi_1^{\ell_3}))) \wedge\Box (\neg(\pi_1^{O}\vee \pi_{1}^{\ell_{4}})) $, which requires robot $1$ to (i) eventually visit locations $\ell_1$, $\ell_2$, and $\ell_{3}$ in this order; and (ii) avoid all a priori unknown obstacles and the region $\ell_{4}$, since, e.g., hostile units may exist there. 
The problem addressed in this paper can be summarized as follows:
\begin{problem} \label{prob}
Given an initial robot configuration $\bbp(0)$ in an environment with unknown geometric structure, the robot dynamics \eqref{eq:rdynamics}, sensors to detect obstacles, and a task specification captured by an LTL formula $\phi$ (which also requires obstacle avoidance), our goal is to generate online, as the robots learn the environment, an infinite sequence of multi-robot actions $\bbu\in\ccalU$ that satisfies $\phi$ and respects the robot dynamics.
\end{problem}

\begin{figure}[t]
  \centering
  \includegraphics[width=1\linewidth]{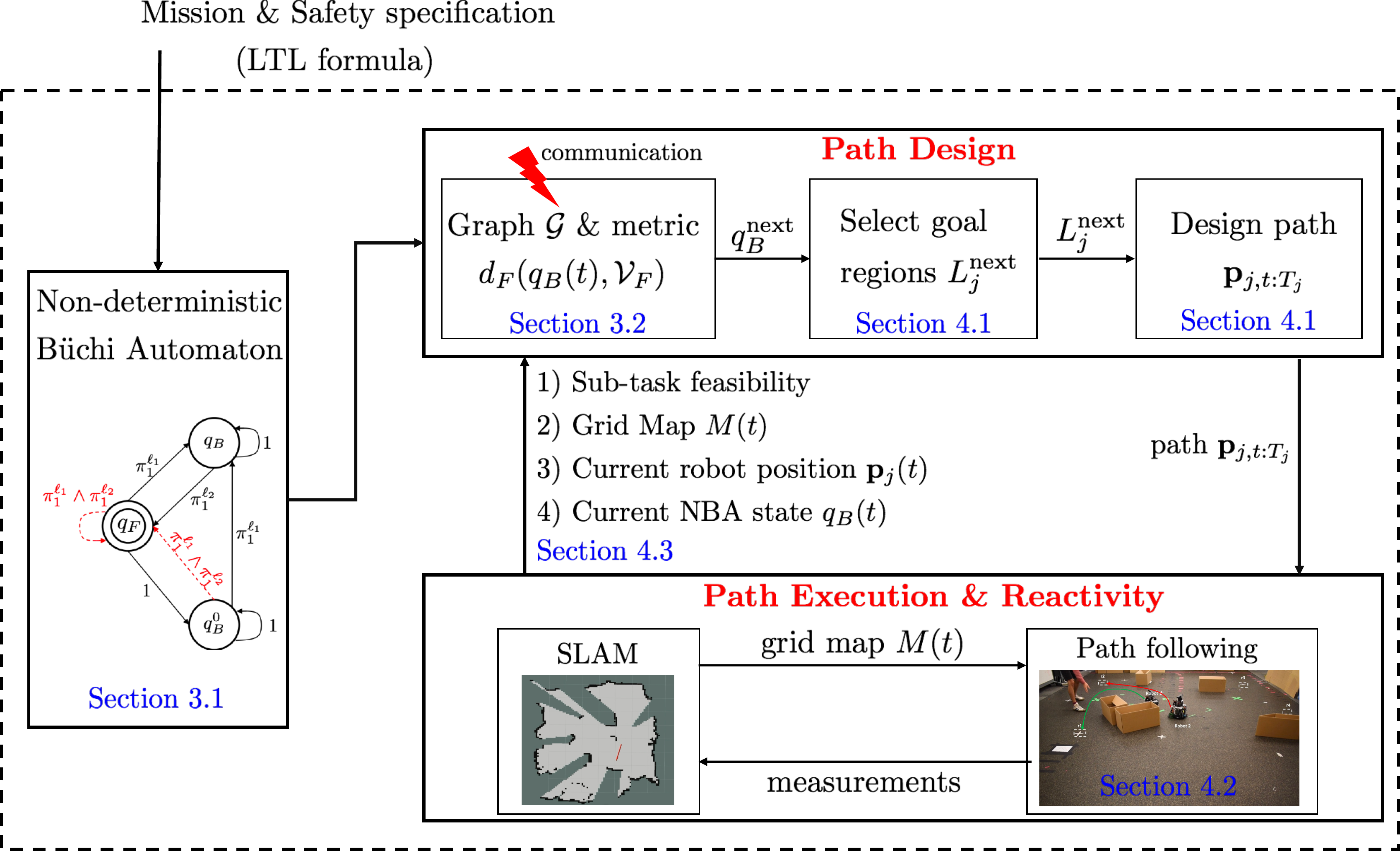}
  \caption{Illustration of the proposed distributed SLAM-based control architecture to solve Problem \ref{prob}. The task is encoded in an LTL formula, translated offline to a Buchi automaton (Section \ref{sec:nba}). This automaton is used to construct a metric measuring how far the robots are from accomplishing the assigned specification (Section \ref{sec:dist}). This metric is used during execution time to decompose the global LTL task into local single-agent reachability problems  in a previously unexplored environment (Section \ref{sec:paths}). Construction of the local reachability tasks requires communication among robots; however, each robot $j$ locally designs paths ${\bf{p}}_{j,t,T_j}$ to accomplish their assigned subtasks (Section \ref{sec:paths}). The robots follow the designed paths while collecting sensor measurements that are used to learn online the map of the environment via a SLAM method (Section \ref{sec:exec}). As the environmental map is continuously learned, the robots may need to revise their paths or design new reachability tasks depending on the feasibility of the already assigned sub-tasks (Section \ref{sec:replan}).}
  \label{fig:solution}
\end{figure}

Problem \ref{prob} can be solved by applying existing abstraction-free temporal logic methods - see, e.g., \cite{vasile2013sampling,luo2019abstraction} - at runtime to continuously update the robot paths as the map of the environment is continuously learned. These methods require searching over a large product automaton space, a task that is computationally challenging to perform repetitively at runtime. To mitigate this challenge, we propose a new reactive LTL planning algorithm that does not rely on searching over a product automaton space. In particular, in Section \ref{sec:aut}, we decompose the temporal logic task into a sequence of local point-to-point navigation problems which allows for efficient and fast re-planning. These navigation problems are formulated and solved online while adapting planning to the continuously learned map of the environment; see Sec. \ref{sec:planning} and Fig. \ref{fig:solution}.

\section{Decomposing Global Temporal Logic Tasks to Local Reachability Tasks}
\label{sec:aut}
%
In this section, we provide an algorithm to decompose global LTL tasks into local single-agent reachability tasks. The proposed method processes an automaton that corresponds to the assigned LTL formula and constructs a metric that measures how far the robots are from accomplishing the assigned task. This algorithm is executed \textit{offline} i.e., before deploying the robots in the unknown environment. Specifically, in Section \ref{sec:nba} we translate the LTL formula into a Non-deterministic B$\ddot{\text{u}}$chi Automaton (NBA). In Section \ref{sec:dist}, we define a distance metric over this automaton that measures how far the robots are from fulfilling the assigned mission. This metric will be used \textit{online} to design single-agent reachability tasks and to guide and adapt planning to the continuously learned environmental map; see Section \ref{sec:planning}. The detailed construction of this metric is presented in Appendix \ref{appA}.

\subsection{From LTL formulas to Automata}\label{sec:nba}

First, we translate the specification $\phi$, constructed using a set of atomic predicates $\mathcal{AP}$, into a Non-deterministic B$\ddot{\text{u}}$chi Automaton (NBA), defined as follows \cite{baier2008principles}. 

\begin{definition}[NBA]
A Non-deterministic B$\ddot{\text{u}}$chi Automaton (NBA) $B$ over $\Sigma=2^{\mathcal{AP}}$ is defined as a tuple $B=\left(\ccalQ_{B}, \ccalQ_{B}^0,\delta_B, \ccalQ_F\right)$, where (i) $\ccalQ_{B}$ is the set of states;
(ii) $\ccalQ_{B}^0\subseteq\ccalQ_{B}$ is a set of initial states; (iii) $\delta_B:\ccalQ_B\times\Sigma\rightarrow2^{\ccalQ_B}$ is a non-deterministic transition relation, and $\ccalQ_F\subseteq\ccalQ_{B}$ is a set of accepting/final states. 
\end{definition}

To interpret a temporal logic formula over the trajectories of the robot system, we use a labeling function $L:\mathbb{R}^{n\times N}\times M(t)\rightarrow 2^{\mathcal{AP}}$ that determines which atomic propositions are true given the current multi-robot state $\bbp(t)\in\mathbb{R}^{n\times N}$ and the current map $M(t)$ of the environment. An infinite-length discrete plan $\tau=\bbp(0)\bbp(1)\dots$ satisfies $\phi$, denoted by $\tau\models\phi$, if it satisfies the accepting condition of the NBA \cite{baier2008principles}. Particularly, if the sequence of observations/symbols in $\sigma=L(\bbp(0), M(0))L(\bbp(1),M(1))\dots$ can generate at least one infinite sequence of NBA states, starting from the initial NBA state, that includes at least one of the final states infinitely often, then we say that $\tau$ satisfies $\phi$. Hereafter, for simplicity we replace $L(\bbp(t),M(t))$ with $L(\bbp(t))$. More details about the NBA accepting condition are provided in Appendix \ref{appA}.


\subsection{Distance Metric Over the NBA}\label{sec:dist}
Following the steps discussed in Appendix \ref{appA}, (i) we prune the NBA by removing infeasible transitions that can never be enabled as they require robots to be in more than one region simultaneously, and (ii) among all feasible transitions we identify which transitions are decomposable, i.e., they can be enabled if the robots solve local/independent reachability-avoidance planning problems. In this section, given such decomposable and feasible NBA transitions we define a function to compute how far an NBA state is from the set of final states. 
This metric will be used in Section \ref{sec:planning} to guide planning towards NBA states that are closer to the final states so that final states are visited infinitely, i.e., the accepting condition of the NBA is satisfied.
To this end, we first construct a directed graph $\ccalG=\{\ccalV,\ccalE\}$, where $\ccalV\subseteq\ccalQ_B$ is the set of nodes and $\ccalE\subseteq\ccalV\times\ccalV$ is the set of edges. As it will be discussed in Appendix \ref{appA}, $\ccalV$ collects all NBA states that can be reached from the initial states by following a sequence of feasible and decomposable transitions. Also, an edge from $q_B\in\ccalV$ to $q_B'\in\ccalV$ exists if the corresponding NBA transition is decomposable and feasible. As it will be discussed in the Appendix \ref{appA}, the transition from $q_B$ to $q_B'$ may correspond to a multi-hop NBA transition, i.e., there may exist intermediate NBA states that need to be visited to reach $q_B'$ from $q_B$. Hereafter, with slight abuse of notation, we denote any multi-hop NBA transition from $q_B$ to $q_B'$ (not necessarily decomposable) by $\delta_B^m(q_B,q_B')$. 

Given the graph $\ccalG$, we define the following distance metric.

\begin{definition}[Distance Metric]
Let $\ccalG=\{\ccalV,\ccalE\}$ be the directed graph that corresponds to the NBA $B$ constructed as discussed in Appendix \ref{appA}. Then, we define the distance function $d: \ccalV \times \ccalV \rightarrow \mathbb{N}$ as follows
\begin{equation}\label{eq:dist}
d(q_B,q_B')=\left\{
                \begin{array}{ll}
                  |SP_{q_B,q_B'}|, \mbox{if $SP_{q_B,q_B'}$ exists,}\\
                  \infty, ~~~~~~~~~\mbox{otherwise},
                \end{array}
              \right.
\end{equation}
where $SP_{q_B,q_B'}$ denotes the shortest path (in terms of hops) in $\ccalG$ from $q_B$ to $q_B'$ and $|SP_{q_B,q_B'}|$ stands for its cost (number of hops). 
\end{definition}

In words, $d:\ccalV\times\ccalV\rightarrow \mathbb{N}$ returns the minimum number of edges in the graph $\ccalG$ that are required to reach a state $q_B'\in\ccalV$ starting from a state $q_B\in\ccalV$. This metric can be computed using available shortest path algorithms, such the Dijkstra method with worst-case complexity $O(|\ccalE| + |\ccalV|\log|\ccalV|)$,
%

%
Next, we define the final/accepting edges in $\ccalG$ as follows.
\begin{definition}[Final/Accepting Edges]\label{def:accEdges}
An edge $(q_B,q_B')\in\ccalE$ is called final or accepting if the corresponding multi-hop NBA transition $\delta_B^m(q_B,q_B')$ includes at least one final state $q_F\in\ccalQ_F$.
\end{definition}

Based on the definition of accepting edges, we define the set $\ccalV_F\subseteq \ccalV$ that collects all states $q_B\in\ccalV$ from which an accepting edge originates, i.e.,
\begin{equation}\label{eq:accNode}
    \ccalV_F = \{q_B\in\ccalV~|~\exists~ \text{accepting edge}~(q_B,q_B')\in\ccalE \}.
\end{equation}
By definition of the accepting condition of the NBA, we have that if at least one of the accepting edges is traversed infinitely often, then the corresponding LTL formula is satisfied. 

Similar to \cite{bisoffi2018hybrid}, we define the distance of any state $q_B\in\ccalV$ to the set $\ccalV_F\subseteq \ccalV$ as
\begin{equation}\label{eq:distF}
d_F(q_B,\ccalV_F)=\min_{q_B'\in\ccalV_F}d(q_B,q_B'),
\end{equation}
where $d(q_B,q_B')$ is defined in \eqref{eq:dist} and $\ccalV_F$ is defined in \eqref{eq:accNode}.

{\bf{Definitions:}} In what follows we introduce definitions associated with multi-hop NBA transitions $\delta_B^m(q_B,q_B')$; see also Example \ref{ex:exFeasSets}. These definitions will be used in Section \ref{sec:planning}. As it will be discussed in Appendix \ref{appA}, the transition from $q_B$ to $q_B'$ is enabled if a Boolean formula denoted by $b^{q_B,q_B'}$ (see \eqref{eq:b}) is satisfied. Given $b^{q_B,q_B'}$ we define the set $\Sigma^{q_B,q_B'}$ that collects all feasible symbols $\sigma^{q_B,q_B'}$ that satisfy $b^{q_B,q_B'}$, i.e., $\sigma^{q_B,q_B'}\models b^{q_B,q_B'}$. As it will be formally defined in Appendix \ref{appA}, a symbol $\sigma^{q_B,q_B'}$ is feasible if it can be generated without requiring any robot to be present in more than one disjoint region simultaneously. Moreover, given $b^{q_B,q_B'}$ we define the set $\ccalN^{q_B,q_B'}\subseteq\ccalN$ that collects the indices of all robots that appear in $b^{q_B,q_B'}$. Also, given any symbol $\sigma^{q_B,q_B'}\in\Sigma^{q_B,q_B'}$ that satisfies $b^{q_B,q_B'}$, we define the set $\ccalR^{q_B,q_B'}\subseteq\ccalN^{q_B,q_B'}$ that collects the indices $j$ of the robots that are involved in generating $\sigma^{q_B,q_B'}$. Also, we denote by $L_{j}^{q_B,q_B'}$ the location that robot $j\in\ccalN^{q_B,q_B'}$ should be located to generate $\sigma^{q_B,q_B'}$. Note that for robots $j\in\ccalN^{q_B,q_B'}\setminus\ccalR^{q_B,q_B'}$, $L_{j}^{q_B,q_B'}$ corresponds to any location $\bbq\in\Omega$ where no atomic predicates are satisfied. If $\delta_B^m(q_B,q_B')$ is a decomposable transition, then we collect in the set $\Sigma^{q_B,q_B'}_{\text{dec}}\subseteq\Sigma^{q_B,q_B'}$ all symbols $\sigma^{q_B,q_B'}$ that can enable this transition in a local reachability fashion. 

\begin{ex}
(a) Consider the Boolean formula $b^{q_B,q_B'}=(\pi_j^{\ell_e}\vee\pi_j^{\ell_m})\wedge(\neg \pi_i^{\ell_e})$, then $\Sigma^{q_B,q_B'}$ is defined as $\Sigma^{q_B,q_B'}=\{\pi_j^{\ell_e},\pi_j^{\ell_m}\}$. 
Note that the symbol $\pi_j^{\ell_e}\pi_j^{\ell_m}$ satisfies $b^{q_B,q_B'}$ but it is not included in $\Sigma^{q_B,q_B'}$, since it is not feasible as it requires robot $j$ to be present simultaneously in regions $\ell_e$ and $\ell_m$. Given the Boolean formula $b^{q_B,q_B'}$, we have that $\ccalN^{q_B,q_B'}=\{j,i\}$.
Also, for $\sigma^{q_B,q_B'}=\pi_j^{\ell_e}$, we have that $\ccalR^{q_B,q_B'}=\{j\}$, $L_{j}^{q_B,q_B'}=\ell_e$. (b) Consider the Boolean formula $b^{q_B,q_B'}=\neg \pi_i^{\ell_e}$. In this case $\Sigma^{q_B,q_B'}$ is defined as $\Sigma^{q_B,q_B'}=\{\varnothing\}$, where $\varnothing$ denotes an empty symbol and is generated when a robot is outside all regions of interest and does not hit any obstacles. Also, we have that $\ccalN^{q_B,q_B'}=\{i\}$, $\sigma^{q_B,q_B'}=\varnothing$, $\ccalR^{q_B,q_B'}=\emptyset$, while $L_{i}^{q_B,q_B'}$ corresponds to any obstacle-free location $\bbq$ in the environment, outside the regions of interest, so that no no atomic predicates are satisfied at $\bbq$. 
\label{ex:exFeasSets}
\end{ex}

\section{Reactive Planning in Unknown Occupancy Grid Maps}\label{sec:planning}


In this section, we present a distributed reactive temporal logic algorithm for robots that operate in unknown environments modeled using occupancy grid maps. The proposed algorithm is summarized in Algorithm \ref{alg:RRT} and requires as inputs (i) the graph $\ccalG$, defined in Section \ref{sec:dist}, and (ii) sets $\Sigma_{\text{dec}}^{q_B,q_B'}$, constructed in Appendix \ref{appA}, that collect all symbols that can enable the transition from $q_B\in\ccalV$ to $q_B'\in\ccalV$ in local reachability fashion; see also Fig. \ref{fig:solution}. The main idea of Algorithm \ref{alg:RRT} is that as the robots navigate the unknown environment, they select NBA states that they should visit next so that the distance to the final states, as per \eqref{eq:distF}, decreases over time. Transitions towards the selected NBA states is accomplished by solving local reachability-avoidance planning problems that are adapted to the continuously learned map. 


\subsection{Distributed Construction of Robot Paths}\label{sec:paths}
Let $q_B(t)\in\ccalV$ be the NBA state that the robots have reached after navigating the unknown environment for $t$ time units. 
At time $t=0$, $q_B(t)$ is selected to be the initial NBA state. 
\begin{algorithm}[t]
\caption{Distributed LTL Control in Unknown Occupancy Grid Maps}
\LinesNumbered
\label{alg:RRT}
\KwIn{ (i) Initial robot configuration $\bbp(0)$, (ii) Graph $\ccalG$ and sets $\Sigma_{\text{dec}}^{q_B,q_B'}$, $\forall q_B,q_B'\in\ccalV$}
Initialize $t=0$, $q_B(0)=q_B^{\text{aux}}$\;\label{rrt:t0}
Select the next NBA state $q_B^{\text{next}}$\; 
Select $\sigma^{\text{next}}\in\Sigma_{\text{dec}}^{q_B(t),q_B^{\text{next}}}$\;\label{rrt:selTarget}
Using $\sigma^{\text{next}}$, compute $\ccalN^{\text{next}}$\;\label{rrt:compN}
\While{accepting condition of NBA is not satisfied}{\label{rrt:sat}
%
\For{$j\in\ccalN^{\text{next}}$}{\label{rrt:for1}
Compute paths $\bbp_{j,t:T_j}$ towards $L_j^{\text{next}}$; see \eqref{eq:localProb}\;\label{rrt:compPaths}
Follow paths, re-plan if needed, and update map\;\label{rrt:ExecReplan}
\If{$\bbp_j(t)\in L_j^{\text{next}}$}{\label{rrt:T}
Wait until all other robots $j\in\ccalN^{\text{next}}$ reach their goal regions.\;}\label{rrt:wait}
\If{$\bbp_j(t)\in L_j^{\text{next}}, \forall i\in\ccalN^{\text{next}}$}{\label{rrt:nowait}
Update $q_B(t)=q_B^{\text{next}}$\;\label{rrt:updNBA}
Select a new NBA state $q_B^{\text{next}}$ and  $\sigma^{\text{next}}$\;\label{rrt:updTarget}
Using $\sigma^{\text{next}}$, update $\ccalN^{\text{next}}$\;\label{rrt:updN}
%
}
}
}
\end{algorithm}
\normalsize
Given the current NBA state $\bbq_B(t)$, the robots coordinate to select a new NBA state, denoted by $q_B^{\text{next}}\in\ccalV$ that they should reach next to make progress towards accomplishing their task [line \ref{rrt:selTarget}, Alg. \ref{alg:RRT}]. This state is selected among the neighbors of $q_B(t)$ in the graph $\ccalG$ based on the following two cases.
If $q_B(t)\notin \ccalV_F$, where $\ccalV_F$ is defined in \eqref{eq:accNode}, then among all neighboring nodes, we select one that satisfies
\begin{equation}\label{eq:minDist}
d_F(q_B^{\text{next}},\ccalV_F) =  d_F(q_B(t),\ccalV_F)-1,  
\end{equation}
i.e., a state that is one hop closer to the set $\ccalV_F$ than $q_B(t)$ is where $d_F$ is defined in \eqref{eq:distF}.  Under this policy of selecting $q_B^{\text{next}}$, we have that eventually $q_B(t)\in\ccalV_F$; design of robot paths to ensure this property will be discussed next. If $q_B(t)\in\ccalV_F$, then the state $q_B^{\text{next}}$ is selected so that $(q_B(t),q_B^{\text{next}})$ is an accepting edge as per Definition \ref{def:accEdges}. This way we ensure that accepting edges are traversed infinitely often and, therefore, the assigned LTL task is satisfied.

Recall that the transition from $q_B(t)$ to $q_B^{\text{next}}$ is decomposable by construction of $\ccalG$. As a result, there exists a symbol that the robots can generate in a local reachability fashion to reach  $q_B^{\text{next}}$ from $q_B(t)$. The symbols that can enable this transition are collected in a set $\Sigma_{\text{dec}}^{q_B,q_B^{\text{next}}}$ constructed in Appendix \ref{appA}. Among all available symbols in $\Sigma_{\text{dec}}^{q_B,q_B^{\text{next}}}$, the robots select one either randomly or based on any user-specified criterion such as distance required to travel to generate it as long as the locations that the robots should visit to generate this symbol are reachable, i.e., not surrounded by obstacles that have already been discovered. By construction of the set $\Sigma_{\text{dec}}^{q_B,q_B^{\text{next}}}$, we have that once the robots reach $q_B^{\text{next}}$ they will be able to stay in this state as long as they keep generating this symbol; see \eqref{eq:b} in Appendix \ref{appA}.  
With slight abuse of notation, we denote the selected symbol by $\sigma^{\text{next}}$ [line \ref{rrt:selTarget}, Alg. \ref{alg:RRT}]. Also, for simplicity of notation we replace $b^{q_B(t),q_B^{\text{next}}}$ with with $b^{\text{next}}$. 
The same simplification extends to all notation with the same superscripts.


In what follows, we discuss how to design individual robot paths $\bbp_{j,t:T_j}$, for all robots $j$, that should be followed within a time interval $[t,T_j]$ to ensure that eventually $\sigma^{\text{next}}$ is generated. Specifically, given $\sigma^{\text{next}}$, every robot $j\in\ccalN^{\text{next}}$ computes the corresponding location $L_j^{\text{next}}$.\footnote{Recall that for robots $j\in\ccalR^{\text{next}}\subseteq \ccalN^{\text{next}}$, $L_j^{\text{next}}$ corresponds to a certain region of interest while for robots $j\in\ccalN^{\text{next}}\setminus\ccalR^{\text{next}}$, $L_j^{\text{next}}$ corresponds to any location in the workspace where no atomic predicates are satisfied. }
Given $L_j^{\text{next}}$, every robot $j\in\ccalN^{\text{next}}$ locally solves the feasibility problem \eqref{eq:localProb} to design its respective path $\bbp_{j,t:T_j}$ towards $L_j^{\text{next}}$ while all other robots stay put at their current locations [lines \ref{rrt:for1}-\ref{rrt:compPaths}, Alg. \ref{alg:RRT}].

\begin{subequations}
\label{eq:localProb}
\begin{align}
& \text{Compute:~}{T_j, \bbu_{j,t:T_j}}~\text{such that}  \label{ObjLoc2}\\
& \ \ \ \ \ \ \ \ \ \ \  \bbp_j(t') \notin \cup_i\{r_i\}\setminus (R_j^t\cup L_j^{\text{next}}), \forall t'\in[t,T_j] \label{constrLoc1}\\
& \ \ \ \ \ \ \ \ \ \ \  \bbp_j(t') \notin \hat{O}(t), \forall t'\in[t,T_j] \label{constrLocObs}\\
& \ \ \ \ \ \ \ \ \ \ \   \bbp_j(T_j) \in L_j^{\text{next}} \label{constrLoc2}\\
%
& \ \ \ \ \ \ \ \ \ \ \   \bbp_j(t'+1)=\bbf_j(\bbp_j(t'),\bbu_j(t')), \label{constrLoc3}
\end{align}
\end{subequations}

\noindent In \eqref{constrLoc1}, $R_j^t$ denotes the region that robot $j$ lies in at time $t$. The first constraint \eqref{constrLoc1} requires that robot $j$ should not visit any region $r_i$ excluding $R_j^t$ and the target region $L_j^{\text{next}}$, for all $t'\in[t,T_j]$. If at time $t$ robot $j$ does not lie in any of the regions $r_i$, then $R_j^t=\emptyset$. This ensures that the multi-robot system can always remain in $q_B(t)$ as the robots $j\in\ccalN^{\text{next}}$ navigate the workspace; see Lemma \ref{lem:binv}. 
Similarly, the second constraint in \eqref{constrLocObs} requires robot $j$ to avoid all obstacles, collected in $\hat{O}(t)$, that are known up to the time instant $t$. The third constraint in \eqref{constrLoc2} requires robot $j$ to be in region $L_j^{\text{next}}$ at a time instant $T_j\geq t$. 
This is required to eventually generate the symbol $\sigma^{\text{next}}$. The last constraint in \eqref{constrLoc3} captures the robot dynamics. Note that among all paths that satisfy these constraints, the optimal one, as a per a user-specified motion cost function, can be selected. 
%
Observe that \eqref{eq:localProb} is a local reachability problem, and, therefore, existing point-to-point navigation algorithms can be used to solve it, e.g., \cite{karaman2011sampling}. Finally observe that given $L_j^{\text{next}}$, robot $j$ locally solves \eqref{eq:localProb} without communicating with any other robot. 

\subsection{Asynchronous Execution of Paths}\label{sec:exec} 
The robots $j\in\ccalN^{\text{next}}$ follow their paths $\bbp_{j,t:T_j}$ to eventually reach their goal regions $L_j^{\text{next}}$. By construction of these paths, the robots may not arrive at the same discrete time instant at their goal regions. To account for this, when a robot $j$ has traversed its path and has reached its goal region, it waits until
all other robots $e\in\ccalN^{\text{next}}$ reach their respective goal regions $L_e^{\text{next}}$ as well [line \ref{rrt:wait}, Alg. \ref{alg:RRT}]. When this happens, the symbol $\sigma^{\text{next}}$ is generated and the NBA state $q_B^{\text{next}}$ is reached as it will be shown in Section \ref{sec:complOpt}. Then a new NBA state is selected [lines \ref{rrt:nowait}-\ref{rrt:updN}, Alg. \ref{alg:RRT}]. This process is repeated indefinitely so that a final state is reached infinitely often [line \ref{rrt:sat}, Alg. \ref{alg:RRT}]. Note also that the robots can execute their paths asynchronously in the sense that the (continuous) time required to move from $\bbp_j(t)$ to $\bbp_j(t+1)$ may differ across the robots without affecting correctness of the proposed algorithm; see Sec. \ref{sec:complOpt}.

\subsection{Reacting to Map Uncertainty}\label{sec:replan}

As the robots follow their paths $\bbp_{j,t:T_j}$, the map $M(t)$ is continuously updated using existing SLAM methods and, therefore, the robot paths need to be accordingly revised [line \ref{rrt:ExecReplan}, Alg. \ref{alg:RRT}]; see also Fig. \ref{fig:solution}. In fact, reactivity to the environment is required in two cases. First, as the robots follow their paths $\bbp_{j,t:T_j}$, they may discover new obstacles. If the discovered obstacles intersect with their paths $\bbp_{j,t:T_j}$, for some $j\in\ccalR^{\text{next}}$, then these robots $j$ locally re-solve \eqref{eq:localProb} using the updated map to design new paths. Note that existing reactive planning algorithms that can address reachability navigation problems in the presence of unknown obstacles can also be used to solve \eqref{eq:localProb}; see e.g., \cite{karaman2011sampling,guzzi2013human,otte2016rrtx,bansal2019combining,ryll2019efficient,elhafsi2019map,zhai2019path,tordesillas2020faster}.

Second, as the the map $M(t)$ is constructed, goal regions $L_j^{\text{next}}$ may turn out to be unreachable because of the structure of the environment (e.g., regions of interest may be surrounded by obstacles). In this case, robots $j$ that cannot reach their corresponding target regions communicate with all other robots so that a new symbol $\sigma^{\text{next}}$ is selected. This process is repeated until a symbol $\sigma^{\text{next}}$ is reached that corresponds to reachable regions $L_j^{\text{next}}$.  
If such a symbol $\sigma^{\text{next}}$ does not exist, then this means that the transition from $q_B(t)$ to $q_B^{\text{next}}$ is infeasible, given the current multi-robot state $\bbp(t)$, due to the a-priori unknown structure of the environment. Then, the graph $\ccalG$ is updated by removing this transition and a new NBA state $q_B^{\text{next}}$ is selected. If there do not exist any candidate NBA states to be selected as $q_B^{\text{next}}$ (see Section \ref{sec:paths}), then the assigned task cannot be accomplished given the current multi-robot and NBA state. Note that this does not necessarily mean that Problem \ref{prob} is infeasible. 
In fact, there may exist another sequence of NBA states for which Problem \ref{prob} is always feasible. 
Conditions under which \eqref{eq:localProb} (and, consequently, Problem \ref{prob}) is always feasible (i.e., paths towards selected goal regions always exist) and, therefore, Algorithm \ref{alg:RRT} is complete, are provided in Section \ref{sec:complOpt}.

\begin{rem}[Communication]
Execution of the paths $\bbp_{j,t:T_j}$ does not require all-time communication among robots, which is typically the case in relevant temporal logic algorithms. Instead, communication is intermittent and is required only in the following two cases: (i) all robots need to communicate with each other to select and update the symbol $\sigma^{\text{next}}$ (ii) every robot $j\in\ccalN^{\text{next}}$ needs to send a message to all other robots in $\ccalN^{\text{next}}$ once it has reached region $L_{j}^{\text{next}}$. 
\end{rem}


\begin{rem}[Conservativeness]\label{rem:conserv}
Recall that once the robots reach a state $q_B(t)$, a new state $q_B^{\text{next}}$ is selected to be reached next.
This state is selected so that the transition from $q_B(t)$ to $q_B^{\text{next}}$ is decomposable and, therefore, it can be enabled by solving local reachability tasks as per \eqref{eq:localProb}. Note that there may be states that can be reached from $q_B(t)$ but not in a local-reachability fashion (see e.g., Figure \ref{fig:conserv} in the Appendix). As a result, such states may not be selected and, therefore, a solution to Problem \ref{prob} may not be found even if it exists. Conditions under which Algorithm \ref{alg:RRT} is complete are discussed in Section \ref{sec:complOpt}.
\end{rem}


\section{Algorithm Analysis}\label{sec:complOpt}


In this section, we examine correctness of the proposed reactive algorithm; all proofs can be found in Appendix \ref{sec:prop}. First, recall from Algorithm \ref{alg:RRT} that once the robots reach a state $q_B(t)\in\ccalV$ a new state $q_B^{\text{next}}\in\ccalV$ is selected that should be reached next.
%
%
By construction of the graph $\ccalG$, the transition from $q_B(t)$ to $q_B^{\text{next}}$ is decomposable i.e., it can be enabled by solving local reachability tasks. As discussed in Remark \ref{rem:conserv}, focusing only on decomposable transitions is conservative, in the sense that there may be NBA states that can be reached from $q_B(t)$ but not in a local-reachability nature; see the example in Figure \ref{fig:conserv} in Appendix \ref{appA}.  As a result, such states will never be selected to be $q_B^{\text{next}}$ and, therefore, a solution to Problem \ref{prob} may not be found even if it exists. 
First, we show that there exists a fragment of LTL formulas for which 
all transitions $\delta_B^m(q_B,q_B')$ introduced in Section \ref{sec:dist} are decomposable and can be enabled by solving local reachability problems \eqref{eq:localProb}. To show this we need to make the following assumption. 





\begin{assumption}\label{as:cnf}
Let $b^{q_B,q_B'}$ be a Boolean formula that needs to be satisfied to enable the NBA transition $q_B'\in\delta_B(q_B,\cdot)$. Assume that $b^{q_B,q_B'}$ is written in conjunctive normal form (CNF). Also, assume that all literals in any given clause of $b^{q_B,q_B'}$ are associated with only one robot, which is not necessarily the same across all clauses in $b^{q_B,q_B'}$.
\end{assumption}

\begin{lem}[From Global LTL to Local Tasks]\label{lem:fragmentLTL}
%
Assume that there exists a symbol $\sigma^{q_B,q_B'}$ that satisfies the corresponding Boolean formula $b^{q_B,q_B'}$ defined in \eqref{eq:b} associated with a multi-hop transition $\delta_B^m(q_B,q_B')$. 
Under Assumption \ref{as:cnf}, the transition $\delta_B^{m}(q_B,q_B')$ is always decomposable, as per Definition \ref{def:feas}, unless this transition requires robots to instantaneously jump from one region of interest to another one which is impossible to achieve given that the regions of interest are disjoint.
%
\end{lem}

In other words, according to Lemma \ref{lem:fragmentLTL}, if there are two NBA states $q_B, q_B'$, where  $q_B'$ can be reached from $q_B$ using a feasible word and without requiring robots to instantaneously move from one region of interest to another one and Assumption \ref{as:cnf} holds for this NBA transition, then this transition can be enabled by solving local reachability tasks; see Example \ref{ex:localTasks}.

Next, we show that if the robots navigate the workspace as per Section \ref{sec:exec}, then the goal NBA state $q_B^{\text{next}}$ will be reached. This result will be used to show that Algorithm \ref{alg:RRT} is complete.

\begin{lem}[Execution of paths]\label{lem:binv}
Assume that at time $t$ the robots reach a state $q_B(t)\in\ccalV$ and design paths $\bbp_{j,t:T_j}$ to reach  $q_B^{\text{next}}\in\ccalV$.
If the robots execute and update the paths $\bbp_{j,t:T_j}$, as per Section \ref{sec:exec}, and eventually they reach their goal regions $L_j^{\text{next}}$ then the transition from $q_B$ to $q_B^{\text{next}}$ will be activated. 
\end{lem}

In what follows, we provide conditions that ensure that Algorithm \ref{alg:RRT} is complete, i.e., if there exists a solution to Problem \ref{prob}, then a solution will be found.

\begin{prop}[(Completeness)]\label{thm:completeness}
Assume that 
(a) the environmental structure and the robot dynamics \eqref{eq:rdynamics} ensure that given any $q_B\in\ccalV$ there exists at least one state $q_B^{\text{next}}\in\ccalV$ satisfying the conditions described in Section \ref{sec:paths} that can be reached, and
%
(b) all robots are equipped with omnidirectional and perfect/deterministic sensors. Under assumption (a)-(b), Algorithm \ref{alg:RRT} is complete, i.e., if there exists a solution to Problem \ref{prob}, then Alg. \ref{alg:RRT} will find it.
\end{prop}


Finally, in the next proposition we show that under assumption (a) of Proposition \ref{thm:completeness}, a solution to Problem \ref{prob} always exists.

\begin{prop}[Existence of Solution]\label{thm:exist}
If Assumption (a) of Proposition \ref{thm:completeness} holds, then there exists a solution to Problem \ref{prob}.
\end{prop}

\begin{rem}[From Global LTL Tasks to Local Tasks]
To check if an LTL formula $\phi$ can be decomposed into local tasks it suffices to convert $\phi$ into a NBA and then manually check if each Boolean formula $b^{q_B,q_B'}$ satisfies Assumption \ref{as:cnf}. Alternatively, it suffices to convert $\phi$ into the graph $\ccalG$ and then check if there is a sequence of edges, starting from the initial state of this graph, denoted by $q_B^{\text{aux}}$ in Appendix \ref{appA}, so that at least one accepting edge is traversed infinitely often. The latter holds since, by construction of the graph $\ccalG$, an edge in $\ccalG$ between two NBA states exists if the corresponding NBA transition is decomposable; see Appendix \ref{appA}.
\end{rem}

\begin{ex}[From Global LTL to Local Tasks]\label{ex:localTasks}
For instance, consider the Boolean formula  $b^{q_B,q_B'}=(\pi_1^{\ell_1}\vee\pi_2^{\ell_1})\wedge (\pi_1^{\ell_2}\vee\pi_2^{\ell_2})$ that is expressed in CNF form. This formula has two clauses: (i) $(\pi_1^{\ell_1}\vee\pi_2^{\ell_1})$ and (ii) $(\pi_1^{\ell_2}\vee\pi_2^{\ell_2})$ The first clause has two literals $\pi_1^{\ell_1}$ and $\pi_2^{\ell_1}$ that are not associated with the same robot. The same also holds for the second clause. As a result, this Boolean formula violates Assumption \ref{as:cnf} and, therefore, it cannot be decomposed into local tasks. In fact,  $b^{q_B,q_B'}$ could be decomposed into local tasks where e.g., robot $1$ should go to $\ell_1$ and robot $2$ should go to $\ell_2$. However, this decomposition is formula-specific and requires additional coordination among the robots, i.e., the individual robot tasks cannot be constructed locally/independently. In contrast, consider the Boolean formula $b^{q_B,q_B'}=(\pi_1^{\ell_1}\vee\pi_1^{\ell_2})\wedge (\pi_2^{\ell_1}\vee\pi_2^{\ell_2})$. This formula satisfies Assumption \ref{as:cnf} and, therefore, it can be satisfied by accomplishing subtasks that are locally/independently constructed where e.g., robot $1$ should go to $\ell_1$ and robot $2$ should go to $\ell_2$.
\end{ex}

\section{Numerical Simulations \& Experiments} \label{sec:Sim}


\begin{figure*}[t]
  \centering
      \subfigure[$t=315$]{
    \label{fig:315}
  \includegraphics[width=0.3\linewidth]{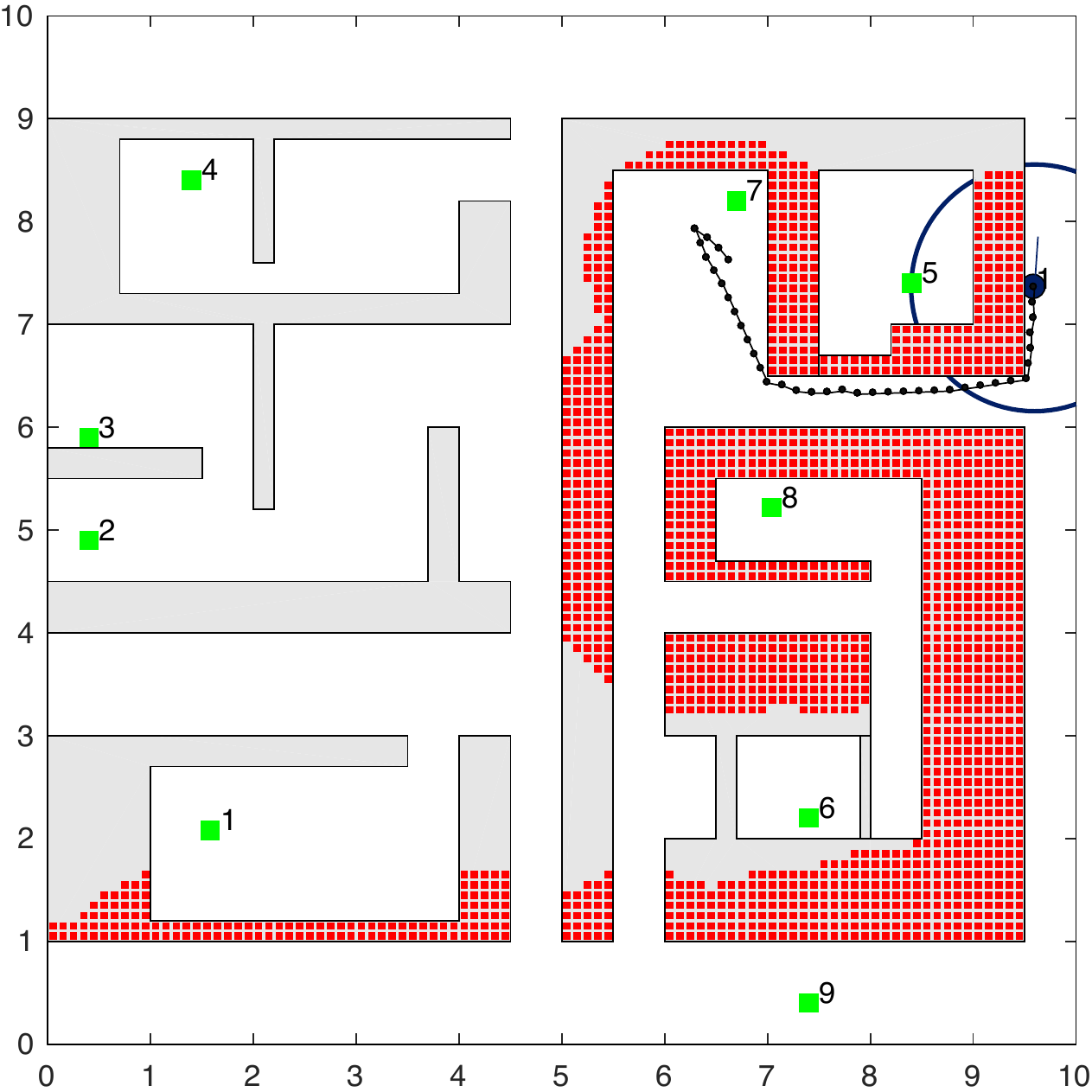}}
  \subfigure[$t=342$]{
    \label{fig:342}
  \includegraphics[width=0.3\linewidth]{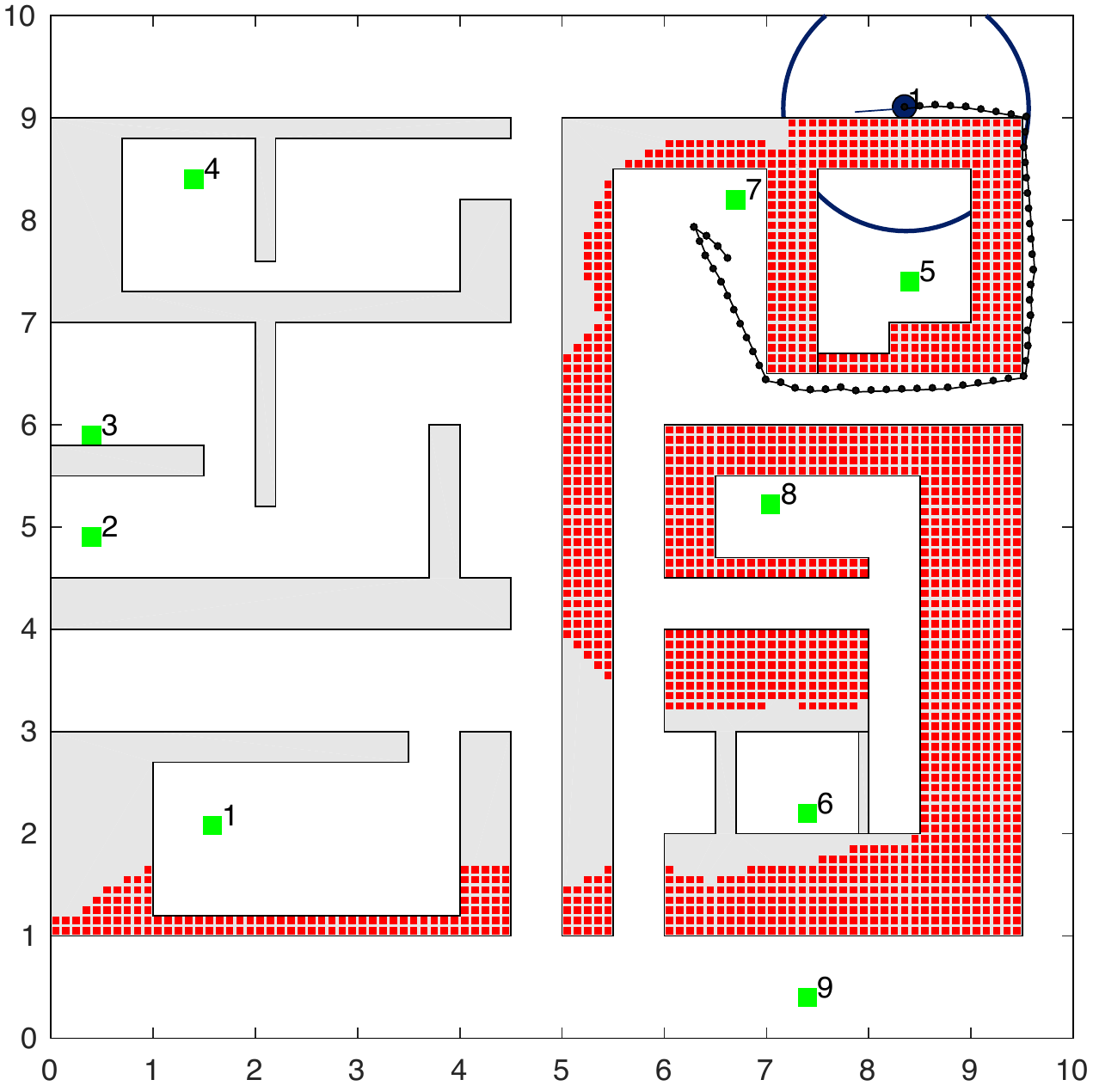}}
    \subfigure[$t=411$]{
    \label{fig:411}
  \includegraphics[width=0.3\linewidth]{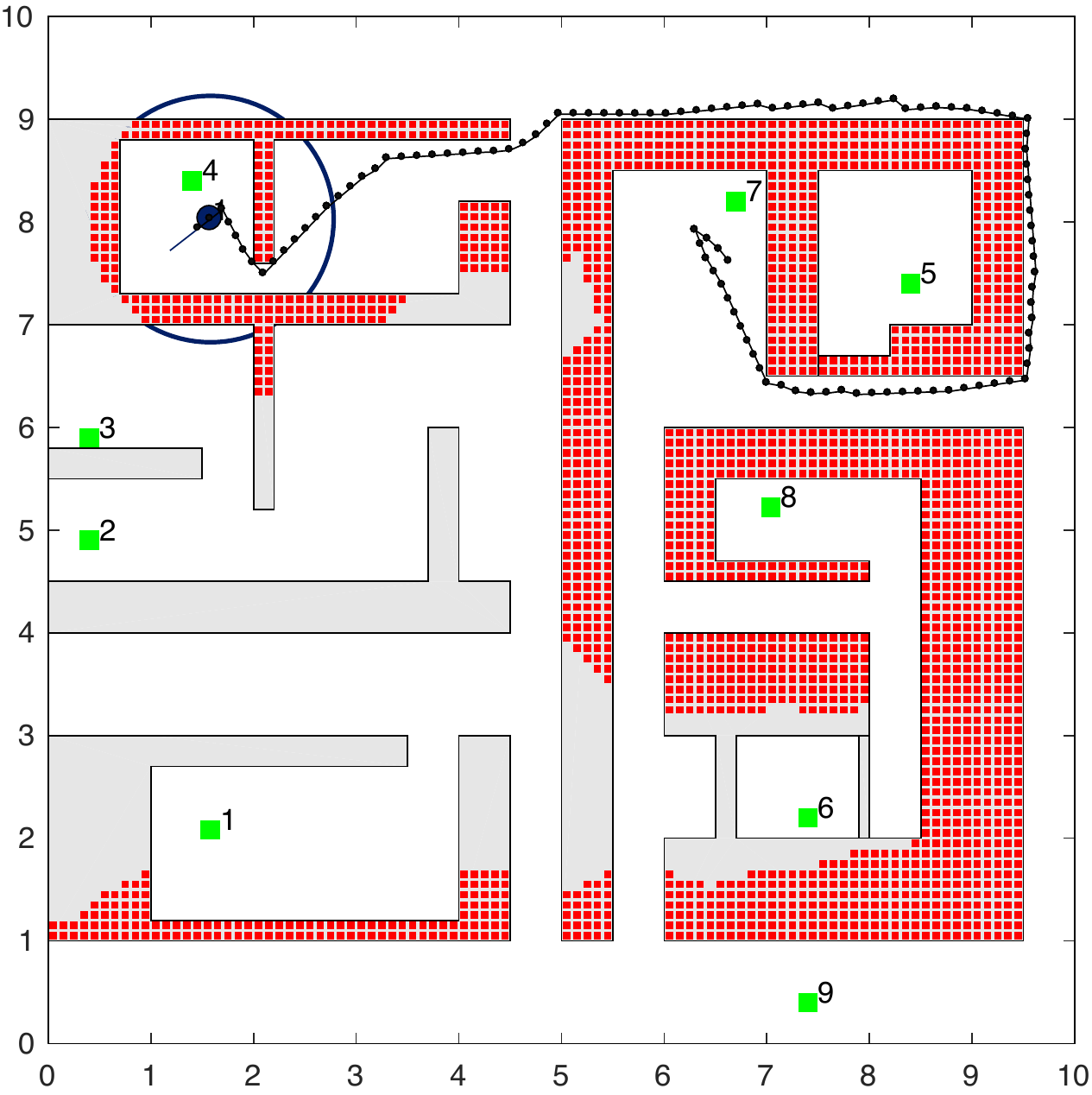}}
  \caption{Case $N=1$, $R=0.3$: Figures \ref{fig:315}-\ref{fig:411} illustrate the robot trajectory while aiming to satisfy $\Diamond(\pi_1^{\ell_5}\vee\pi_1^{\ell_4})$, i.e., to visit either region $\ell_5$ or $\ell_4$. At time instant $t=315$ (Fig. \ref{fig:315}) the robot is heading towards $\ell_5$ and at $t=342$ (Fig. \ref{fig:342}) it realizes, based on the constructed occupancy grid map, that $\ell_5$ is surrounded by obstacles and, therefore, not reachable. As a result, it moves towards $\ell_4$ ((Fig. \ref{fig:411})) and arrives there at $t=411$. The blue circle denotes the sensing range and rhe red squares stands for the grid cells of the map $M(t)$ that are occupied by obstacles. The gray region corresponds to unknown obstacles. 
  }
  \label{fig:env1}
\end{figure*}

\begin{table*}[t]
\caption{Scalability Analysis-Comparative Results}
\label{tab:scale}
\centering
\begin{tabular}{|l|l|l|l|l|l|}
\hline
      & \multicolumn{2}{c|}{Average Re-planning Time (secs)} & \multicolumn{3}{c|}{Algorithm \ref{alg:RRT}} \\ \hline
      $N$   & Algorithm \ref{alg:RRT}        & STyLuS         & Local Map Update (secs)  & $t_F^1$(iteration/time(hrs))  & $t_F^2$(iteration/time(hrs))  \\ \hline
N=1   & 0.05                   & 4.12        & 0.007                                               &    883~/~0.012               & 1065~/~0.015               \\ \hline
N=20  & 0.03                   & 313.69     & 0.007                                               &      952~/~0.17               &   1032~/~0.18                  \\ \hline
N=100 & 0.03                   & 1349.54        &0.009                                               &    1088~/~0.49                 &   1124~/~0.57                  \\ \hline
N=200 & 0.04                   & 2883.52        & 0.01                                              &   1028~/~1.13                  &    1067~/~1.23                 \\ \hline
N=500 & 0.05                   & 9271.57        & 0.008                                              & 1032~/~2.7                    & 1070~/~2.9                     \\ \hline
\end{tabular}
\end{table*}

In this section, we present simulations that illustrate the efficiency and scalability of the proposed algorithm and hardware experiments that demonstrate the robustness of the proposed method under imperfect sensing. Simulations have been implemented using MATLAB R2016b on a computer with Intel Core i7 3.1GHz and 16Gb RAM.\footnote{Videos with simulations and hardware experiments can be found on \url{https://vimeo.com/489973649}.} 


\subsection{Numerical Experiments}\label{sec:sim1}
{\bf{Map-based Planning in Unknown Environments with Unreachable Regions:}} 
In what follows, we consider a single robot with differential drive dynamics equipped with a perfect/deterministic range-limited sensor defined as follows

\begin{equation}\label{eq:sensorSim}
   s(\bbp_j(t),\bbq)=
            \begin{cases}
               1~\text{if~}(\bbq\in O) \wedge (\lVert \bbp_j(t)-\bbq \rVert\leq R),\\
              0~\text{otherwise}.
            \end{cases} 
\end{equation}
with sensing range $R=1.2$. The robot resides in a $10\times10$ environment shown in Figure \ref{fig:env1} and is tasked with a surveillance mission that requires to visit regions of interest eventually or infinitely often
while always avoiding obstacles. This task is captured by the following LTL formula
\begin{equation}\label{task1}
 \phi= (\bigwedge_{e=1}^{2}\square\Diamond\xi_e)\wedge(\bigwedge_{e=3}^6\Diamond\xi_e)\wedge(\neg\xi_3\ccalU\xi_4)\wedge(\square\neg\xi_{\text{obs}})
\end{equation}
where $\xi_e$ is a Boolean formula that requires the robot to visit a specific regions of interest, for all $e\in\{1,\dots,6\}$, and $\xi_{\text{obs}}$ is true if the robot hits any obstacle. In particular, the Boolean formulas $\xi_e$ are defined as: (i) $\xi_1=(\pi_1^{\ell_6}\vee\pi_1^{\ell_3})$; (ii) $\xi_2=\pi_1^{\ell_1}$; (iii) $\xi_3=\pi_1^{\ell_2}$; (iv) $\xi_4=(\pi_1^{\ell_5}\vee\pi_1^{\ell_4})$; (v) $\xi_5=\pi_1^{\ell_8}$; and (vi) $\xi_6=\pi_1^{\ell_9}$. In words, \eqref{task1} requires (a) the Boolean formulas $\xi_1$ and $\xi_2$ to be satisfied infinitely often, (b) the Boolean formulas $\xi_3$, $\xi_4$, $\xi_5$, and $\xi_6$ to be satisfied eventually, (c) never satisfy $\xi_3$ until $\xi_4$ is satisfied, and (d) always avoid obstacles. This formula corresponds to a NBA with $14$ states and $61$ transitions. The corresponding pruned automaton has $37$ transitions while all of them are decomposable; see Definition \ref{def:feas} and Proposition \ref{lem:fragmentLTL}. The runtime to prune this automaton was $0.15$ seconds approximately. 

Figure \ref{fig:env1} illustrates the benefit of using online learned maps for planning in unknown environments. Specifically, incorporating mapping into the control loop allows robots to reason about the environment and make appropriate control decisions to accomplish the assigned task which would be quite challenging using map-free approaches e.g., \cite{guzzi2013human,bansal2019combining}
Particularly, Figure \ref{fig:env1} shows the trajectory that the robot follows to eventually satisfy $\xi_1=(\pi_1^{\ell_5}\vee\pi_1^{\ell_4})$ that requires robot $1$ to visit either location $\ell_5$ or $\ell_4$. Observe in Figure \ref{fig:env1} that $\ell_5$ is surrounded by obstacles and, therefore, it is not accessible. Initially, robot $1$ is moving towards $\ell_5$ and as it builds the map of the environment it realizes that $\ell_5$ is not reachable. When this happens, a new symbol that satisfies $\xi_1$ is selected and, therefore, robot $1$ decides to move towards $\ell_4$. A similar behavior is observed when the Boolean formula $\xi_1=(\pi_1^{\ell_6}\vee\pi_1^{\ell_3})$ needs to be satisfied. 

{\bf{Scalability Analysis:}} Next, we examine the performance of Algorithm \ref{alg:RRT} with respect to the number of robots and we provide comparative simulation studies against state-of-the-art centralized algorithms \cite{kantaros2018text,guo2013revising} that can be used for re-planning as the map of the environment is continuously learned. 

We consider a team of $N\in\{1,20,100,200,500\}$ robots with differential drive dynamics that reside in the environment shown in Figure \ref{fig:env1} while all robots are equipped with omnidirectional sensors of limited range $R=1.2$ as in \eqref{eq:sensorSim}. The robots are tasked with accomplishing the task captured in \eqref{task1}. The difference is that the Boolean formulas $\xi_e$ require collaboration between multiple-robots. For instance, when $N=20$, we have that $\xi_1=\wedge_{j=1}^{15}(\pi_j^{\ell_6}\vee\pi_j^{\ell_3})$ which requires all robots with index $j\in\{1,\dots, 15\}$ to visit either region $\ell_3$ or $\ell_6$ and $\xi_2=\wedge_{j=8}^{20}(\pi_j^{\ell_2}\vee\pi_j^{\ell_3})$ which requires all robots with index $j\in\{8,\dots, 20\}$ to visit either region $\ell_2$ or $\ell_3$. 

Table \ref{tab:scale} summarizes the scalability analysis of the proposed algorithm. Specifically, the first column shows the number of robots, the second column shows the average time required for a robot to design a new path as the map is updated. To design paths we solve \eqref{eq:localProb} using the $\text{A}^*$ algorithm \cite{russell2002artificial}. Specifically, we represent the learned occupancy grid map, which consists of $10,000$ cells, as a graph and we apply $\text{A}^*$ on it to design paths towards the goal regions. Then, we synthesize controllers to follow the designed path. 
Note that the runtime to design new paths ranges between $0.03$ and $0.06$ seconds showing that Algorithm \ref{alg:RRT} can design new paths very fast for large-scale multi-robot systems with hundred of robots due to its decentralized nature. The third collumn shows the average runtime required if existing centralized methods are used for re-planning \cite{guo2013revising,kantaros2018text}. Specifically, \cite{guo2013revising} employs partially known transition systems to model the interaction of the robot and environment which are updated as the environment is discovered. Motion plans are synthesized by composing the transition systems with the NBA giving rise to a product automaton. When the transition system is updated, the product automaton is locally updated and new paths are designed by computing new paths in the product automaton from the current product state to a final product state. To compare Algorithm \ref{alg:RRT} against this approach, we use a graph-based representation of the occupancy grid map as the transition system. Our implementation of \cite{guo2013revising} required more than $1$ hour to design a path over the product automaton for a single-robot system and ran out of memory for case studies with more than one robot. To enhance \cite{guo2013revising}, we integrate it with STyLuS* \cite{kantaros2018text}, a recently proposed temporal logic planning method that scales well with the number of robots. Specifically, we use \cite{kantaros2018text} to design paths, instead of building a product automaton and computing paths over it. The third column in Table \ref{tab:scale} shows the average runtime required for STyLuS* to design new paths. To compute this runtime, we consider cases where $q_B(t)=q_B^0$, i.e., the robots are still in the initial NBA state, and new paths towards a final NBA state need to be computed as in \cite{guo2013revising}. Observe in the third column of Table \ref{tab:scale} that STyLuS* can also handle large-scale multi-robot systems but the runtime required for replanning using Algorithm \ref{alg:RRT} is significantly smaller due to its decentralized nature and the decomposition of the LTL task into `smaller' single-agent tasks.

The fourth column shows the average time required per robot to update the map. Note that the map is trivially updated without the need of sophisticated SLAM methods since deterministic/perfect sensors are considered in \eqref{eq:sensorSim}. The fifth and sixth column present the number of discrete time instants required to visit a final NBA state for the first and the second time respectively. These columns also include the corresponding total runtime required by Alg. \ref{alg:RRT}. Note that Alg. \ref{alg:RRT} has been implemented on a single computer and, therefore, control decisions are made sequentially, instead of in parallel, across the robots which explains why the runtime increases significantly as the number of robots increases.

\begin{figure}[t]
  \centering
  \includegraphics[width=0.8\linewidth]{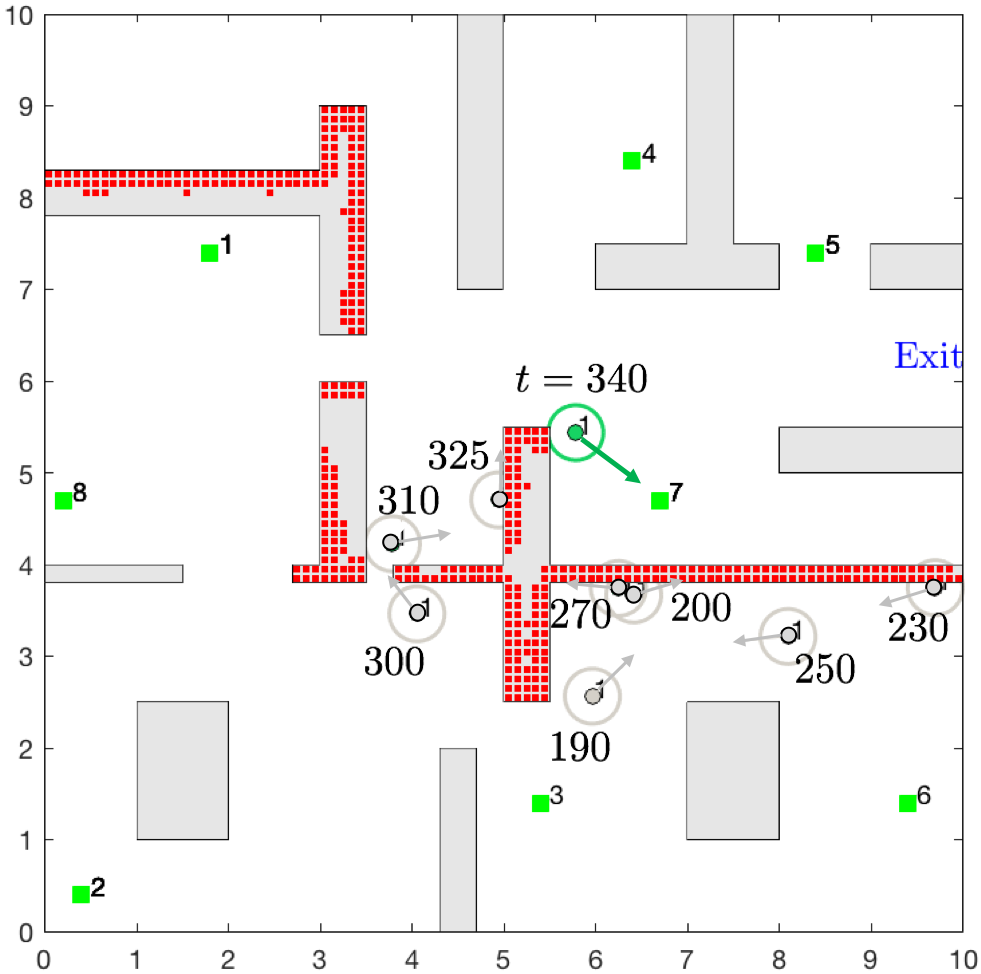}
  \caption{Case $N=1$, $R=0.3$: Figure \ref{fig:340} illustrate the robot position (discs) and actions (arrows) while it is heading towards region $7$. The green circle denotes the sensing range. At time $t=190$ the robot is not aware of the horizontal obstacle that is between it and region $7$. As a result it moves toward the obstacle, until it is fully discovered. Then, a new obstacle-free path is eventually discovered to reach region $7$ ($t=340$). The red squares stands for the grid cells of the map $M(t)$ that are occupied by obstacles. 
  }
  \label{fig:340}
\end{figure}

\begin{figure}[t]
  \centering
  \includegraphics[width=0.8\linewidth]{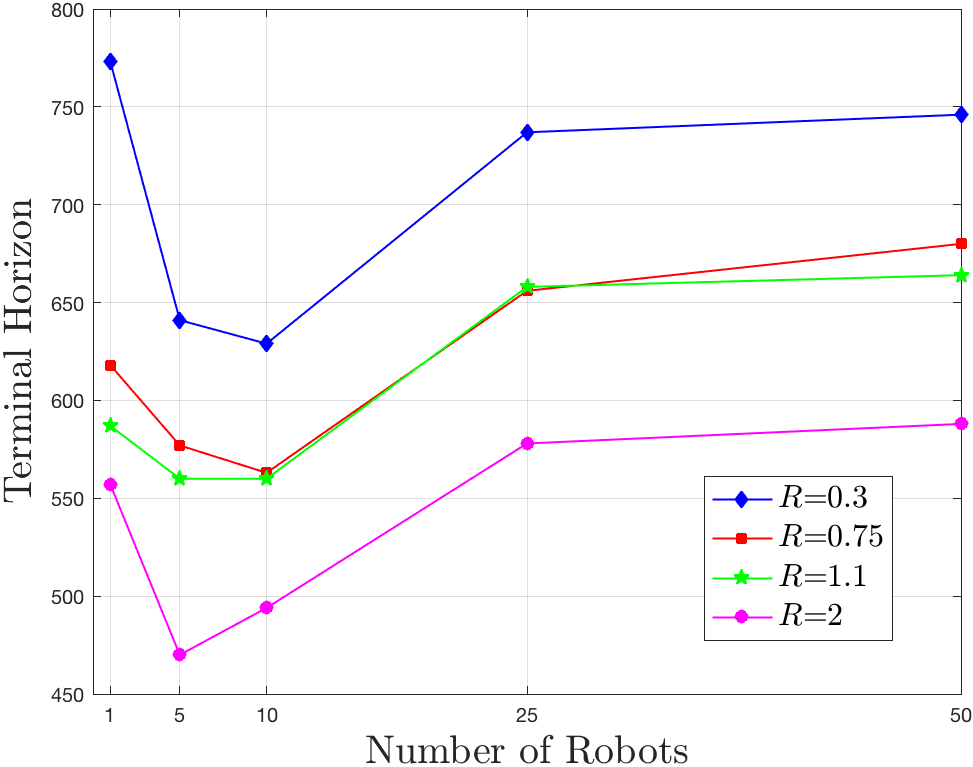}
  \caption{Effect of sensing range. 
  }
 \label{fig:effect}
\end{figure}

{\bf{Effect of Sensing Range:}}  In this set of simulations, we consider a team of $N\in\{1,5,10,25, 50\}$ robots with differential drive dynamics that reside in an unknown $10\times 10$ environment; see Fig. \ref{fig:340}. The robots are equipped with omnidirectional sensors of limited range $R\in\{0.3,0.75,1.1, 2\}$. 
The robots are tasked with providing delivery services to known locations, in a user-specified order, while always avoiding obstacles and dangerous regions. Once and only when services are provided, the robots should exit the building. This task is captured by the following LTL formula
\begin{align}\label{eq:task}
    \phi&=\Diamond(\xi_1\wedge\Diamond(\xi_2\wedge\Diamond(\xi_3\wedge(\Diamond\xi_4 \wedge (\Diamond\xi_5))))) \wedge\nonumber \\&\wedge (\square\neg \xi_6) \wedge (\square\neg \xi_{\text{obs}}) \wedge (\Diamond\square\xi_{\text{exit}})
\end{align}
where $\xi_e$ is a Boolean formula that requires a sub-team of robots to visit a specific region to provide e.g., supplies, for all $e\in\{1,\dots,6\}$, while $\xi_{\text{exit}}$ is true if all robots visit the exit of the building. For instance, when $N=1$, the Boolean formulas $\xi_e$ are defined as: (i) $\xi_1=(\pi_1^8\vee\pi_1^5)$; (ii) $\xi_2=\pi_1^7$; (iii) $\xi_3=\pi_1^4$; (iv) $\xi_4=(\pi_1^1\vee\pi_1^2)$; (v) $\xi_5=\pi_1^6$;  and (vi) $\square\neg \xi_6$ requires the robot to always  keep a distance of at least $0.3$ from region $3$. This LTL formula corresponds to a NBA with $17$ states $103$ transitions while the pruned automaton has $43$ transitions.
Figure \ref{fig:340} depicts the trajectory that a robot follows to reach $\ell_7$ as it learns the occupancy grid map through a sensor with a range $R=0.3$. Figure \ref{fig:effect} shows that as the sensing range $R$ increases, the total number of discrete time instants required to visit a final state for the first time decreases, since the larger the sensing range, the faster the robots can learn the map of the environment.  

\subsection{Hardware Experiments}
The proposed reactive algorithm was also evaluated experimentally to test its efficiency in real-time re-planning and reactivity. Specifically, Alg. \ref{alg:RRT} was implemented in MatLab with ROS and run on the Scarab platform developed at the University of Pennsylvania \cite{michael2008experimental}. The Scarab is a differential drive robot equipped with a Hokuyo UTM-30LX scanning laser that can obtain range measurements in a $270^0$ field-of-view up to a distance of $30$m and an Intel RealSense camera with a $180^0$ field-of-view. Laser-based sensing is used for robot localization and for building an occupancy grid map using the gmapping ROS package implementing the SLAM approach presented in \cite{grisetti2007improved}. Note this sensor violates the assumptions made in Proposition \ref{thm:completeness} as it is neither perfect nor omnidirectional.  

\begin{figure}[t]
    \centering
    \includegraphics[width=0.95
    \linewidth]{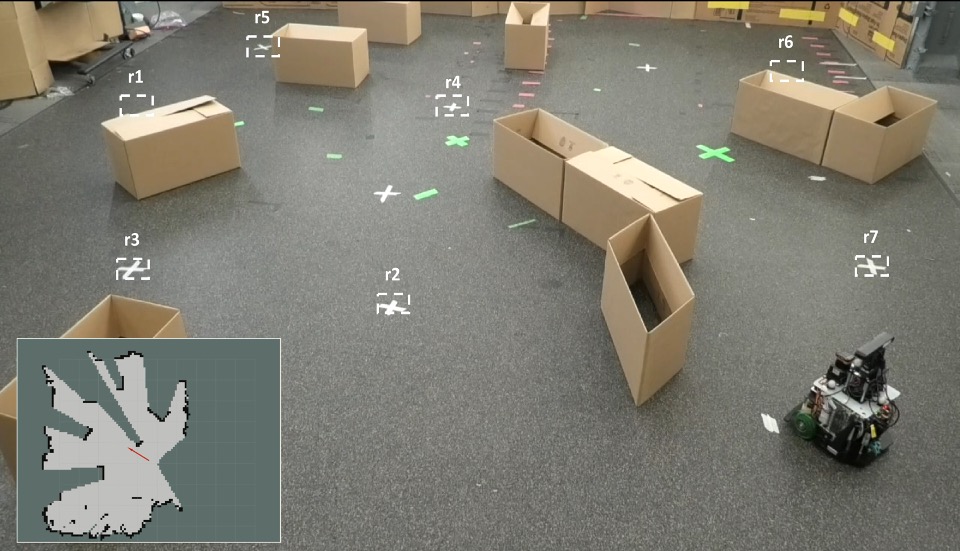}
    \caption{Experiment setup: A differential drive robot equipped with a LiDAR sensor navigates indoor environments with unknown and static obstacles.}
    \label{fig:exp}
\end{figure}
{\bf{LiDAR-based \& Map-based Planning:}} First, we validated Alg. \ref{alg:RRT} in two complex indoor environments with \textit{unknown} and \textit{static} obstacles shown in Figure \ref{fig:exp}. The robot is responsible for accomplishing a surveillance task captured by the following LTL formula: 
\begin{align}\label{task3}
    \phi=&\Diamond(\pi_1^{\ell_4})\wedge\Diamond(\pi_1^{r_1})\wedge\Diamond(\pi_1^{r_2}\vee\pi_1^{r_5})\wedge\Diamond(\pi_1^{r_6}\vee\pi_1^{r_4})\nonumber\\& \wedge\Diamond(\pi_1^{r_7}\vee\pi_1^{r_5})\wedge (\neg \pi_1^{r_1}\ccalU \pi_1^{r_3})\wedge(\square \neg \xi^{\text{obs}}),
\end{align}
The task in \eqref{task3} requires the robot to eventually visit the locations $\ell_4$, $\ell_1$, either $\ell_2$ or $\ell_5$, either $\ell_6$ or $\ell_4$, either $\ell_7$ or $\ell_5$ and never visit $\ell_1$ until $\ell_3$ is visited while always avoiding the a priori unknown obstacles. Observe in Figure \ref{fig:exp} that obstacles can be both convex and non-convex. 
To ensure that the robot paths never go too close to the obstacles, paths are designed as per \eqref{eq:localProb} after inflating the obstacles. As anticipated, the few failures we recorded were associated with the inability of the SLAM approach to accurately localize the robot in long missions.

{\bf{LiDAR-based \& Map-Free Planning:}} 
In the second set of experiments we consider a team of two robots that have to accomplish the following task
\begin{align}\label{task4}
    \phi=&\Diamond(\pi_2^{\ell_2})\wedge\Diamond(\pi_1^{\ell_1}\wedge\Diamond(\pi_1^{\ell_3}\wedge\pi_2^{\ell_3}))\nonumber\\&\wedge\Diamond\square(\pi_1^{\ell_4})\wedge\Diamond\square(\pi_2^{\ell_4})\wedge(\square \neg \xi^{\text{obs}}).
\end{align}
The LTL task in \eqref{task4} requires robot $1$ and robot $2$ to eventually visit locations $\ell_1$ and $\ell_2$, respectively, e.g., to collect data. Then the robots should meet in $\ell_3$ e.g., to exchange their collected data or upload it to a data center, and then go to the exit $\ell_4$ of the room and stay there forever. 

\begin{figure*}[t]
  \centering
      \subfigure[Initial state]{
    \label{fig:a1}
  \includegraphics[width=0.32\linewidth]{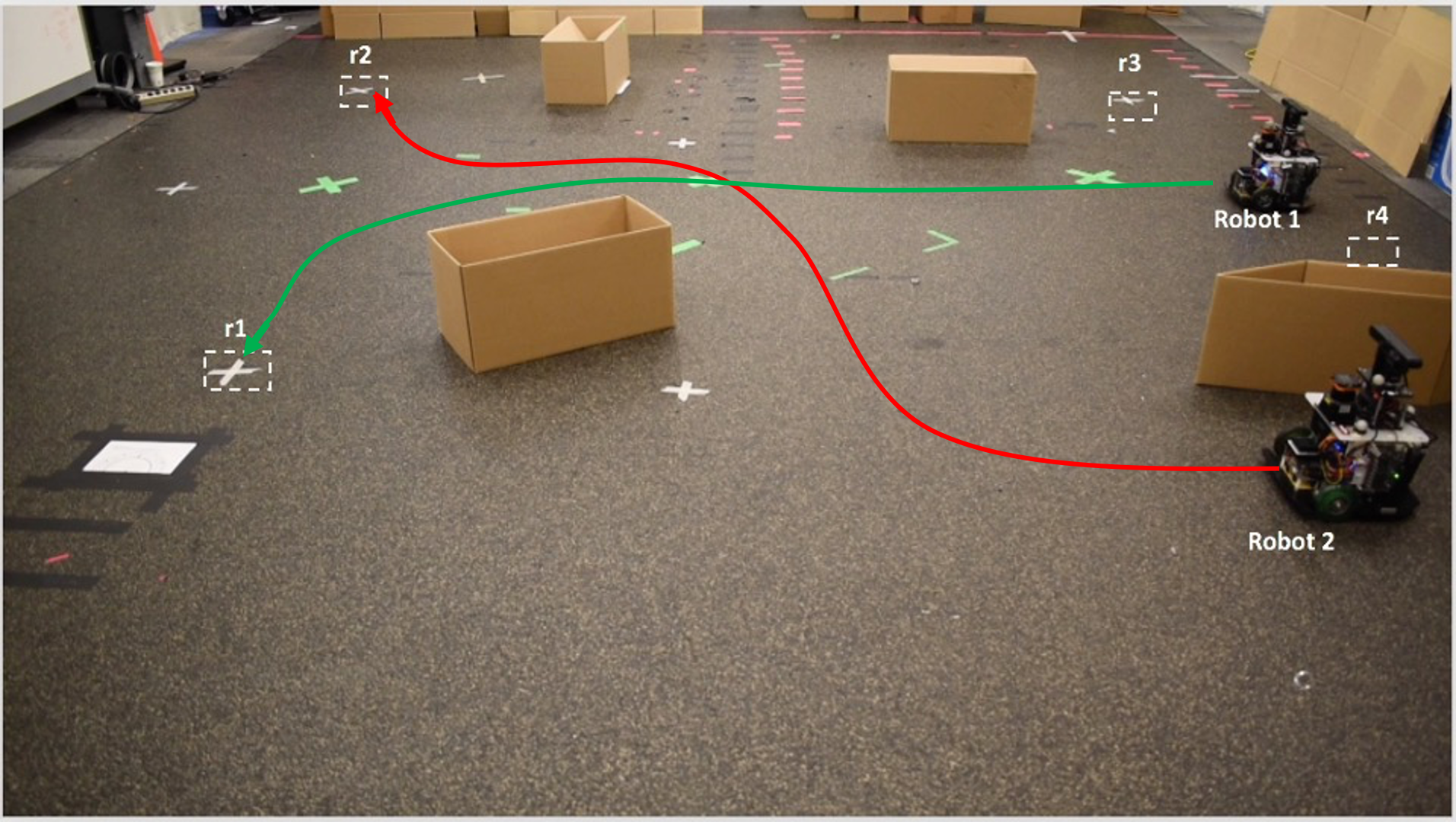}}
  \subfigure[Unexpected obstacle]{
    \label{fig:b1}
  \includegraphics[width=0.32\linewidth]{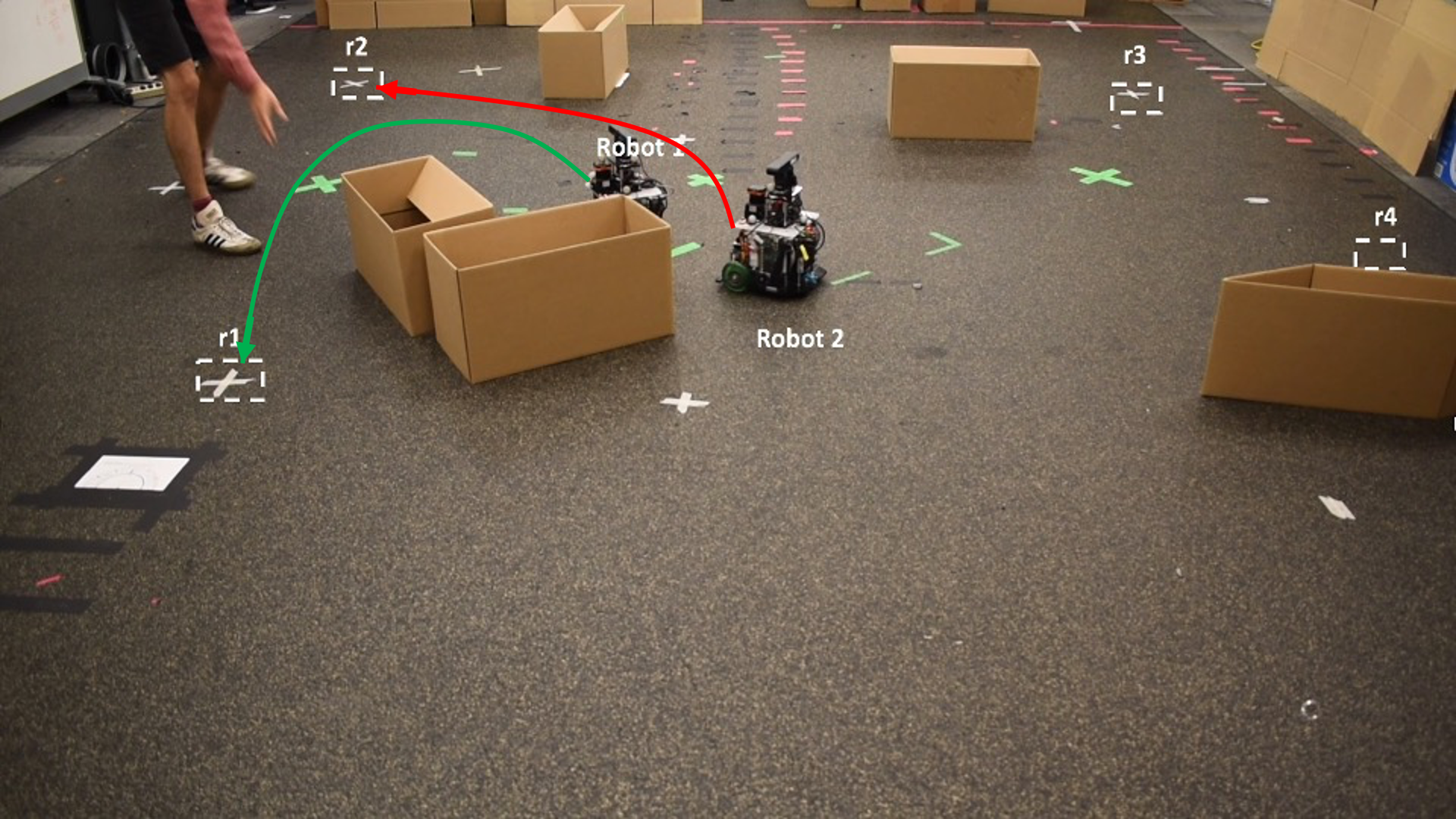}}
      \subfigure[Reactive Planning]{
    \label{fig:d1}
  \includegraphics[width=0.32\linewidth]{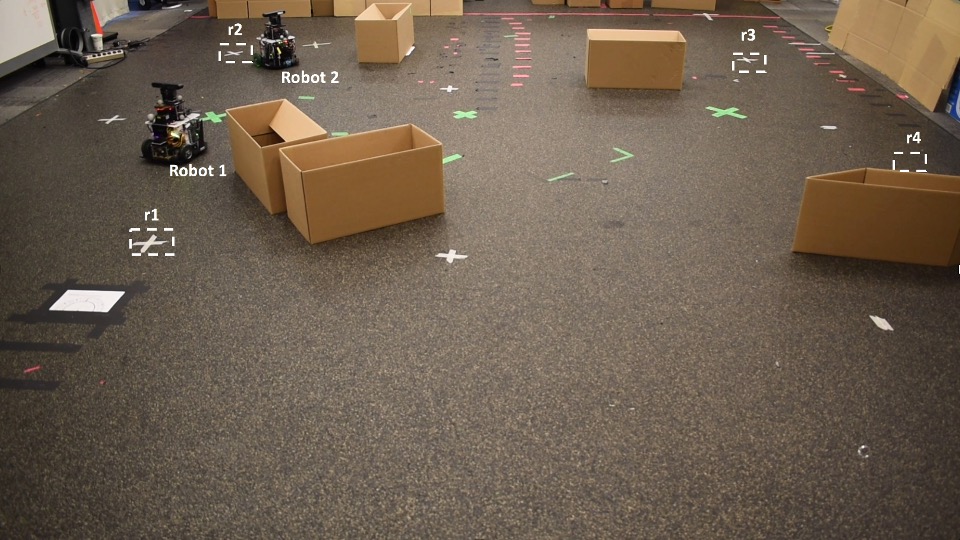}}
  \caption{Camera-based planning in unknown environments: Robot $1$ and $2$ are heading towards locations $1$ and $2$ respectively. Obstacles are placed at run-time in the workspace. The robots adapt planning to these unexpected changes in the environment using sensor feedback.
  }
  \label{fig:expMulti1}
\end{figure*}

In this set of experiments, to solve \eqref{eq:localProb}, i.e., to design reactive robot paths towards regions of interest, we rely on a map-free approach proposed in \cite{guzzi2013human}. Specifically, \cite{guzzi2013human} proposes a heuristic approach to avoid dynamic obstacles such as other mobile robots or humans. Using sensor information the heading of dynamic obstacles is estimated which is then used to compute the robot velocity so that all visible obstacles and robots are avoided within a time horizon. Figure \ref{fig:expMulti1} 
shows a case where the robots react to unexpected obstacles that are placed in front of them at runtime. We would like to emphasize again that Algorithm \ref{alg:RRT} allows the robots to react to unknown structure of the environment quickly which would be computationally challenging using available centralized temporal logic planning methods.

Note that \cite{guzzi2013human} is map-free, i.e., it does not rely on building maps; instead navigation is accomplished using sensor feedback. This allows to navigate dynamic environments as opposed to the purely map-based approach considered in the previous set of experiments.
Nevertheless, map-based approaches allows the robots to reason about the structure of the environment and make appropriate control decisions to accomplish their tasks as discussed in \ref{sec:sim1}. 

\section{Conclusion} \label{sec:Concl}
In this paper, we proposed a new reactive, distributed, and abstraction-free temporal logic planning approach for multi-robot systems that reside in unknown environments modeled as occupancy grid maps. The robots are tasked with accomplishing complex tasks, captured by temporal logic formulas. The robots are equipped with sensors that allow them to continuously learn and update a map of the unknown environment while adapting mission planning to it. Theoretical guarantees along with numerical simulations and hardware experiments support the proposed framework.

%
\appendices
\section{Decomposition of Global LTL Tasks Into Local Reachability Tasks} \label{appA} 

In this section, we provide an algorithm to decompose global LTL tasks into local reachability tasks. The proposed method processes an automaton that corresponds to the assigned LTL formula and constructs a metric that measures how far the robots are from accomplishing the assigned task. This algorithm is executed \textit{offline} i.e., before deploying the robots in the unknown environment. Specifically, first we translate the LTL formula into a Non-deterministic B$\ddot{\text{u}}$chi Automaton (NBA), as discussed in Section \ref{sec:nba}. For convenience, we provide the definition of the NBA again and we provide a formal definition of the accepting condition of the NBA. In Section \ref{sec:prune} we prune this automaton by removing infeasible transitions i.e., transitions that can be activated only if a robot is present in more than one location simultaneously. Then, in Section \ref{sec:dec} we provide conditions under which feasible NBA transitions can be enabled by solving local reachability tasks. Such transitions are called decomposable and are used to define the graph $\ccalG$ and the distance metric over it in Section \ref{sec:dist}. 

\subsection{From LTL formulas to Automata}\label{sec:nbaR}

First, we translate the specification $\phi$, constructed using a set of atomic predicates $\mathcal{AP}$, into a Non-deterministic B$\ddot{\text{u}}$chi Automaton (NBA), defined as follows \cite{baier2008principles}. 

\begin{definition}[NBA]
A Non-deterministic B$\ddot{\text{u}}$chi Automaton (NBA) $B$ over $\Sigma=2^{\mathcal{AP}}$ is defined as a tuple $B=\left(\ccalQ_{B}, \ccalQ_{B}^0,\delta_B, \ccalQ_F\right)$, where (i) $\ccalQ_{B}$ is the set of states;
(ii) $\ccalQ_{B}^0\subseteq\ccalQ_{B}$ is the set of initial states; (iii) $\delta_B:\ccalQ_B\times\Sigma\rightarrow2^{\ccalQ_B}$ is a non-deterministic transition relation, and $\ccalQ_F\subseteq\ccalQ_{B}$ is a set of accepting/final states. 
\end{definition}

To interpret a temporal logic formula over the trajectories of the robot system, we use a labeling function $L:\mathbb{R}^{n\times N}\times M(t)\rightarrow 2^{\mathcal{AP}}$ that determines which atomic propositions are true given the current multi-robot state $\bbp(t)\in\mathbb{R}^{n\times N}$ and the current map $M(t)$ of the environment. An infinite-length discrete plan $\tau=\bbp(0)\bbp(1)\dots$ satisfies $\phi$, denoted by $\tau\models\phi$, if the word $\sigma=L(\bbp(0), M(0))L(\bbp(1),M(1))\dots$ yields an accepting NBA run defined as follows. First, a run of $\rho_B$ of $B$ over an infinite word $\sigma=\sigma(1)\sigma(2)\dots\sigma(t)\dots\in(2^{\mathcal{AP}})^{\omega}$, is a sequence $\rho_B=q_B(0)q_B(1)q_B(2)\dots,q_B(t),\dots$, where $q_B(0)\in\ccalQ_B^0$ and $q_B(t+1)\in\delta_B(q_B(t),\sigma(t))$, $\forall t\in\mathbb{N}$. 
A run $\rho_B$ is called \textit{accepting} if at least one final state appears infinitely often in it. In words, an infinite-length discrete plan $\tau$ satisfies an LTL formula $\phi$ if it can generate at least one accepting NBA run. Hereafter, for simplicity we replace $L(\bbp(t),M(t))$ with $L(\bbp(t))$.
\subsection{Pruning Infeasible NBA Transitions}\label{sec:prune}
%
In what follows, we define the \textit{feasible/infeasible} NBA transitions $q_B'\in\delta_B(q_B,\cdot)$. Transitions that are infeasible are removed giving rise to a pruned automaton. To formally define the \textit{feasible} NBA transitions we first need to introduce the following definitions. 
%


\begin{definition}[Feasible symbols $\sigma_j\in\Sigma_j$]\label{defn:infTSi}
A symbol $\sigma_j\in\Sigma_j:=2^{\mathcal{AP}_j}$ is \textit{feasible} if and only if $\sigma_j\not\models b_j^{\text{inf}}$, where $b_j^{\text{inf}}$ is a Boolean formula defined as 
\begin{equation}\label{bInfTSi}
b_j^{\text{inf}}=\vee_{\forall r_i}( \vee_{\forall r_e}(\pi_j^{r_i}\wedge\pi_j^{r_e})).
\end{equation}
where $b_j^{\text{inf}}$ requires robot $j$ to be present simultaneously in more than one disjoint region.
\end{definition}


Note that the Boolean formula $b_j^{\text{inf}}$ is satisfied by any finite symbol $\sigma_j\in\Sigma_j$ that requires robot $j$ to be present in two or more disjoint regions, simultaneously. For instance, the symbol $\sigma_j=\pi_j^{r_i}\pi_j^{r_e}$ satisfies $b_j^{\text{inf}}$ if $r_i$ and $r_e$ are disjoint regions. 
Next, we define the feasible symbols $\sigma\in\Sigma$.

\begin{definition}[Feasible symbols $\sigma\in\Sigma$]\label{defn:infPTS}
A symbol $\sigma\in\Sigma$ is \textit{feasible} if and only if $\sigma_j\not\models b_j^{\text{inf}}$, for all robots $j$,
where $\sigma_j=\Pi|_{\Sigma_j}\sigma$, $b_j^{\text{inf}}$ is defined in \eqref{bInfTSi}, and $\Pi|_{\Sigma_j}\sigma$ stands for the projection of the symbol $\sigma$ onto $\Sigma_j$.\footnote{For instance, $\Pi|_{\Sigma_j}(\pi_j^{r_e}\pi_m^{r_h})=\pi_j^{r_e}$.} We denote by $\Sigma_{\text{feas}}\subseteq\Sigma$ the set of all feasible symbols $\sigma\in\Sigma$. 
\end{definition}

\begin{algorithm}[t]
\caption{Computing Decomposable Transitions}
\LinesNumbered
\label{alg:prune}
\KwIn{Pruned NBA}
\KwOut{Graph $\ccalG=\{\ccalV,\ccalE\}$, $\Sigma^{q_B,q_B'}_{\text{dec}}$, $\forall (q_B,q_B')\in\ccalE$}
Add auxiliary states and transitions\label{prune:aux}\;
Compute the set $\ccalD_{q_B^{\text{aux}}}$\;\label{prune:reachAux}
Construct set of nodes $\ccalV=\ccalD_{q_B^{\text{aux}}}$ and initialize set of edges $\ccalE=\emptyset$\;\label{prune:initG}
\For{$q_B\in\ccalD_{q_B^{\text{aux}}}$}{
    Initialize $\ccalQ_{q_B}^{\text{next}}=\emptyset$\;
    \For{each run \eqref{eq:run}}{\label{prune:forRun}
        \If{$\delta_B^m(q_B,q_B')$ is decomposable (Def. \ref{def:feas})}{\label{prune:ifDec}
         Collect in $\Sigma^{q_B,q_B'}_{\text{dec}}$ all symbols $\sigma^{q_B,q_B'}$ that satisfy Def. \ref{def:feas} \;\label{prune:upd2}
         $\ccalE=\ccalE\cup\{(q_B,q_B')\}$\label{prune:updEdges}
        }}
    }
Define graph $\ccalG=\{\ccalV,\ccalE\}$ \;\label{prune:graph}
\end{algorithm}

Next, we define the sets that collect all feasible symbols that enable a transition from an NBA state $q_B$ to another, not necessarily different, NBA state $q_B'$. 
This definition relies on the fact that transition from a state $q_B$ to a state $q_B'$ is enabled if a Boolean formula, denoted by $b^{q_B,q_B'}$ and defined over the set of atomic predicates $\mathcal{AP}$, is satisfied.  

\begin{definition}[Sets of Feasible Symbols]
Let $\mathcal{AP}_{q_B,q_B'}$ be a set that collects all atomic predicates that appear in the Boolean formula $b^{q_B,q_B'}$ associated with an NBA transition $ q_B'\in\delta_B(q_B,\cdot)$. Given $b^{q_B,q_B'}$, we define the set $\Sigma^{q_B,q_B'}\subseteq 2^{\mathcal{AP}_{q_B,q_B'}}$ 
that collects all symbols $\sigma$ that (i) satisfy the corresponding Boolean formula $b^{q_B,q_B'}$, i.e., $\sigma\models b^{q_B,q_B'}$; and (ii) $\sigma\in\Sigma_{\text{feas}}$. 
\label{defn:feasSet}
\end{definition}

%
%

Given the sets $\Sigma^{q_B,q_B'}$, we define the infeasible NBA transitions defined as follows:

\begin{definition}[Infeasible NBA Transitions]\label{def:feas1}
A transition $q_B'\delta_B(q_B,\cdot)$ is infeasible if $\Sigma^{q_B,q_B'}=\emptyset$.
\end{definition}

\begin{figure}[t]
  \centering
      \subfigure[Original NBA]{
    \label{fig:origNBA}
  \includegraphics[width=0.46\linewidth]{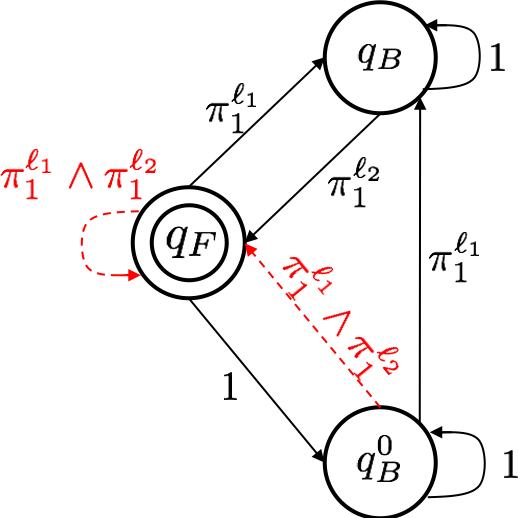}}
  \subfigure[Pruned NBA (Def. \ref{def:feas1})]{
    \label{fig:feas1}
  \includegraphics[width=0.38\linewidth]{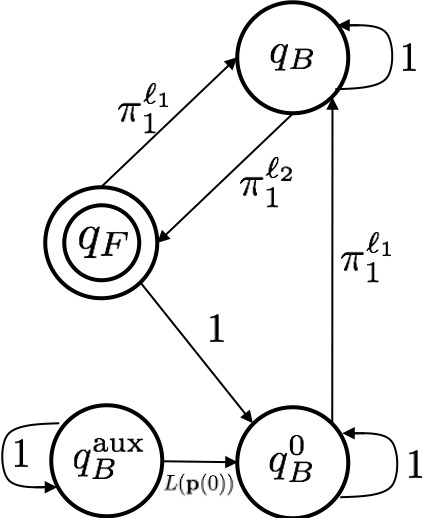}}
%
  \caption{Graphical illustration of Algorithm \ref{alg:prune} applied to the NBA corresponding to $\phi=\square\Diamond(\pi_1^{\ell_1})\wedge\square\Diamond(\pi_1^{\ell_2})$. This formula requires robot $1$ to visit infinitely often $\ell_1$ and $\ell_2$. Figure \ref{fig:origNBA} shows the original NBA where the states $q_B^0$ and $q_F$ correspond to the initial and final state of the NBA. The red dashed transitions are infeasible, as per Definition \ref{def:feas}, as they require robot $1$ to be in $\ell_1$ and $\ell_2$ simultaneously.  These transitions are removed  yielding the pruned NBA shown in Figure \ref{fig:feas1}. An auxiliary state $q_B^{\text{aux}}$ is added to the pruned NBA; see Figure \ref{fig:feas1}.
  Observe in Figure \ref{fig:feas1} that $\ccalD_{q_B^{\text{aux}}}=\{q_B^{\text{aux}},q_B^0,q_B\}$. Also, there are two runs in the form \eqref{eq:run} originating from $q_B$: $\rho_1=q_Bq_B$ and $\rho_2=q_B q_F q_B^0$. Note that the run $\rho=q_B q_F q_B$ cannot be generated as there is no feasible symbol that satisfies $\pi_1^{\ell_1}\wedge\pi_1^{\ell_2}$. In fact, this run requires robot $1$ to `jump' from $\ell_1$ to $\ell_2$ instantaneously. 
  Thus $\ccalQ_{q_B}^{\text{next}}=\{q_B,q_B^0\}$. Similarly, for $q_B^0$ there are two runs in the form \eqref{eq:run}: $\rho_1=q_B^0q_B^0$ and $\rho_2=q_B^0 q_B$ and $\ccalQ_{q_B^0}^{\text{next}}=\{q_B,q_B^0\}$.
  %
  }
  \label{fig:prune}
\end{figure}

In words, an NBA transition is infeasible if it is enabled only when at least one robot is present in more than one location in the environment simultaneously. All infeasible NBA transitions are pruned; see e.g., Figure \ref{fig:prune}. 


\begin{figure}
    \centering
       \subfigure[Original NBA]{
    \label{fig:origNBAU}
  \includegraphics[width=0.45\linewidth]{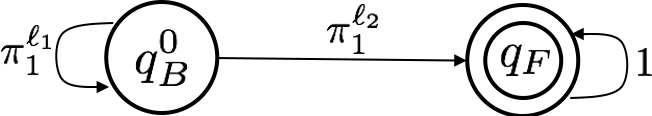}}
    \subfigure[Pruned NBA]{
    \label{fig:finalNBA2}
  \includegraphics[width=0.45\linewidth]{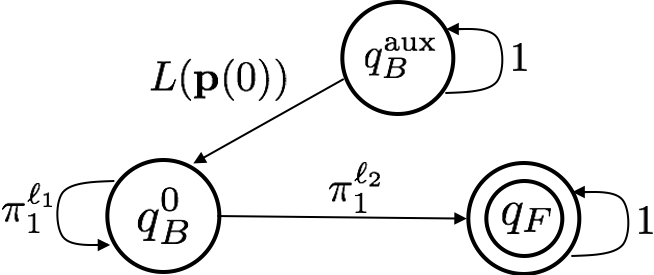}}
    \caption{Effect of the initial robot state in computing decomposable transitions. Figure \ref{fig:origNBAU} depicts the NBA corresponding to $\phi=\pi_1^{\ell_1}\ccalU\pi_1^{\ell_2}$. All transitions are feasible as per Def. \ref{def:feas1}. Observe in Figure \ref{fig:finalNBA2} that if the initial robot state $\bbp(0)=\ell_1$, then $\ccalD_{q_B^{\text{aux}}}=\{q_B^{\text{aux}},q_B^0\}$. Note that the final state $q_F$ is not reachable from $q_B^0$. Also, $\ccalQ_{q_B^{\text{aux}}}^{\text{next}}=\{q_B^0,q_B^{\text{aux}}\}$ and $\ccalQ_{q_B^0}^{\text{next}}=\{q_B^0\}$. Note that the transition from $q_B^0$ to $q_F$ (red dashed line) does not satisfy Def. \ref{def:feas} and therefore the final state $q_F$ is not reachable. 
    If $\bbp(0)=\ell_2$, then $\ccalD_{q_B^{\text{aux}}}=\{q_B^{\text{aux}},q_F\}$, $\ccalQ_{q_B^{\text{aux}}}^{\text{next}}=\{q_B^{\text{aux}},q_F\}$, and $\ccalQ_{q_{F}}^{\text{next}}=\{q_F\}$. 
    }
    \label{fig:NBA1}
\end{figure}

\subsection{Computing Decomposable NBA Transitions}\label{sec:dec}

In this section, we discuss when the accepting condition of the (pruned) NBA can be satisfied by requiring each robot to solve a sequence of local reachability problems; see Algorithm \ref{alg:prune} and Figures \ref{fig:prune}-\ref{fig:NBA1}.
In what follows, we assume that the NBA has been pruned by removing infeasible transitions.


First, to take into account the initial multi-robot state in this process, we introduce an auxiliary state $q_B^{\text{aux}}$ and transitions from $q_B^{\text{aux}}$ to all initial states $q_B^0\in\ccalQ_B^0$ so that $b^{q_B^{\text{aux}},q_B^{\text{aux}}}=1$ and $b^{q_B^{\text{aux}},q_B^{0}}= L(\bbp(0))$, i.e., transition from $q_B^\text{aux}$ to $q_B^0$ can always be enabled based on the initial robot configuration; see also Figure \ref{fig:NBA1} [line \ref{prune:aux}, Alg. \ref{alg:prune}]. Hereafter, the auxiliary state $q_B^{\text{aux}}$ is considered to be the initial state of the resulting NBA. 

Second, we collect all NBA states that can be reached from $q_B^{\text{aux}}$ in a possibly multi-hop fashion, using a finite and feasible word (i.e., a finite sequence of feasible symbols), so that once these states are reached the robots can always remain in them as long as needed using the same symbol that allowed them to reach this state. Formally, let $\ccalD_{q_B^{\text{aux}}}$ be a reachable set that collects all NBA states $q_B$ (i) that have a feasible self-loop, as per Definition \ref{def:feas1}, i.e., $\Sigma^{q_B,q_B}\neq\emptyset$ and (ii) for which there exists a finite and feasible word $w$, i.e., a finite sequence of feasible symbols, so that starting from $q_B^{\text{aux}}$, a finite NBA run $\rho_w$ is incurred that ends in $q_B$ and activates the self-loop of $q_B$ [line \ref{prune:reachAux}, Alg. \ref{alg:prune}]. In math, $\ccalD_{q_B^{\text{aux}}}$ is defined as:
\begin{align}\label{eq:reach}
    \ccalD_{q_B^{\text{aux}}}=&\{q_B\in\ccalQ_B|\\&(\Sigma^{q_B,q_B}\neq\emptyset)\wedge(\exists w~\text{s.t.}~\rho_{w}=q_B^{\text{aux}}\dots \bar{q}_B q_Bq_B)\nonumber\}.
\end{align}
By definition of $q_B^{\text{aux}}$, we have that $q_B^{\text{aux}}\in\ccalD_{q_B^{\text{aux}}}$. Note that the reachable set $\ccalD_{q_B^{\text{aux}}}$ can be computed by viewing the pruned NBA a graph and applying graph search-based methods.

Third, among all possible pairs of states in $\ccalD_{q_B^{\text{aux}}}$, we examine which transitions, possibly multi-hop, can be enabled using feasible symbols so that once these states are reached the robots can always remain in them forever using the same symbol that allowed them to reach this state. Formally, consider any two states $q_B, q_B'\in\ccalD_{q_B^{\text{aux}}}$ (i) that are connected through a - possibly multi-hop - path in the NBA, and (ii) for which there exists a symbol, denoted by $\sigma^{q_B,q_B'}$, so that if it is repeated a finite number of times starting from $q_B$, the following finite run can be generated: 
\begin{equation}\label{eq:run}
    \rho=q_B q_B^1\dots q_B^{K-1}q_B^{K}q_B^{K},
\end{equation}
where $q_B'=q_B^K$, for some finite $K>0$. In \eqref{eq:run}, the run is defined so that (i) $q_B^k\neq q_B^{k+1}$, for all $k\in\{1,K-1\}$; (ii) $q_B^{k}\notin\delta_B(q_B^k,\sigma^{q_B,q_B'})$ for all $\forall k\in\{1,\dots,K-1\}$, i.e., the robots cannot remain in any of the intermediate states (if any) that connect $q_B$ to $q_B'$ either because a self-loop does not exist or because $\sigma^{q_B,q_B'}$ cannot activate this self-loop; and (iii) $q_B'\in\delta_B(q_B',\sigma^{q_B,q_B'})$ i.e., there exists a feasible loop associated with $q_B'$ that is activated by $\sigma^{q_B,q_B'}$. Due to (iii), the robots can remain in $q_B'$ as long as $\sigma^{q_B,q_B'}$ is generated. The fact that the finite repetition of a \textit{single} symbol needs to generate the run \eqref{eq:run} precludes multi-hop transitions from $q_B$ to $q_B'$ that require a robot to jump from one region of interest to another one instantaneously as such transitions are not meaningful as discussed in Section \ref{sec:PF}; see e.g., Fig. \ref{fig:feas1}. Hereafter, with slight abuse of notation we denote the multi-hop transition incurred due to the run \eqref{eq:run} by  $\delta_{B}^m(q_B,q_B')$.

%
The symbol $\sigma^{q_B,q_B'}$ is selected from $\Sigma^{q_B,q_B'}$ (see Definition \ref{defn:feasSet}) that collects all feasible symbols that satisfy the following Boolean formula \eqref{eq:b}:
\begin{equation}\label{eq:b}
   b^{q_B,q_B'}=b^{q_B,q_B^1}\wedge b^{q_B^2,q_B^3}\wedge\dots b^{q_B^{K-1},q_B^{K}}\wedge b^{q_B^{K},q_B^{K}}.  
\end{equation}
In words, the Boolean formula in \eqref{eq:b} is the conjunction of all Boolean formulas $b^{q_B^{k-1},q_B^{k}}$ that need to be satisfied simultaneously to reach $q_B'=q_B^K$ through a given multi-hop path associated with the run \eqref{eq:run}. By definition of \eqref{eq:b}, if there exists a symbol $\sigma^{q_B,q_B'}$ that satisfies $b^{q_B,q_B'}$ then the finite word generated by repeating $K+1$ times the symbol $\sigma^{q_B,q_B'}$ can yield the run in \eqref{eq:run}. 
%
%
%
Note that given a state $q_B$, finding a symbol $\sigma^{q_B,q_B'}$ and a state $q_B'$ can be accomplished by viewing the NBA as a directed graph and applying graph-search methods on it.



Next, we define when the multi-hop transition $\delta_{B}^m(q_B,q_B')$ is decomposable, i.e., when the word $\sigma^{q_B,q_B'}$ can be generated if the robots accomplish local reachability-avoidance tasks. To define this we introduce the following definitions; they were also introduced in Section \ref{sec:aut}. First, given the Boolean formula $b^{q_B,q_B'}$ in \eqref{eq:b}, we define the set $\ccalN^{q_B,q_B'}\subseteq\ccalN$ that collects the indices of all robots that appear in $b^{q_B,q_B'}$. Also, given a symbol $\sigma^{q_B,q_B'}\in\Sigma^{q_B,q_B'}$, we define the set $\ccalR^{q_B,q_B'}\subseteq\ccalN^{q_B,q_B'}$ that collects the indices $j$ of the robots that are involved in generating $\sigma^{q_B,q_B'}$. Also, we denote by $L_{j}^{q_B,q_B'}$ the location that robot $j\in\ccalN^{q_B,q_B'}$ should be located to generate $\sigma^{q_B,q_B'}$. Note that for robots $j\in\ccalN^{q_B,q_B'}\setminus\ccalR^{q_B,q_B'}$, $L_{j}^{q_B,q_B'}$ corresponds to any location $\bbq\in\Omega$ where no atomic predicates are satisfied.

\begin{definition}[Decomposable NBA transitions]\label{def:feas}
A transition $\delta_{B}^m(q_B,q_B')$ corresponding to a run as in \eqref{eq:run} is decomposable if for all symbols  $\sigma^{q_B,q_B}\in\Sigma^{q_B,q_B}$, there exists at least one symbol $\sigma^{q_B,q_B'}\in\Sigma^{q_B,q_B'}$ that satisfies \eqref{eq:b}
such that either (A) $\ccalR^{q_B,q_B}\cap\ccalR^{q_B,q_B'}=\emptyset$ or (B) $L_{j}^{q_B,q_B}=L_{j}^{q_B,q_B'}$ for all $j\in\ccalR^{q_B,q_B}\cap\ccalR^{q_B,q_B'}\neq\emptyset$.
\end{definition}

Observe that if a multi-hop transition $\delta^m_B(q_B,q_B')$ is decomposable, then it can be enabled by solving \textit{local/independent} reachability-avoidance problems due to conditions (A)-(B). Specifically, as discussed in Section \ref{sec:planning}, as soon as the state $q_B$ is reached, then to reach $q_B'$, the robots in $\ccalR^{q_B,q_B}$ suffice to remain idle in their current positions while robots in $\ccalR^{q_B,q_B'}$ should move to a region of interest $L_{j}^{q_B,q_B'}$ while avoiding obstacles and/or other regions of interest. The latter also holds for the robots in $\ccalN^{q_B,q_B'}\setminus\ccalR^{q_B,q_B'}$. As shown in Section \ref{sec:complOpt} this ensures that eventually the symbol $\sigma^{q_B,q_B'}$  will be generated and the transition $\delta^m_B(q_B,q_B')$ will be enabled. 
%
%
Note that there may be more than one symbol $\sigma^{q_B,q_B'}$ that satisfies Definition \ref{def:feas} and can enable the decomposable transition from $q_B$ to $q_B'$.  Hereafter, we collect all these symbols in the set $\Sigma_{\text{dec}}^{q_B,q_B'}\subseteq \Sigma^{q_B,q_B'}$ [line \ref{prune:upd2}, Alg. \ref{alg:prune}]. 

\begin{figure}
    \centering
  \includegraphics[width=0.7\linewidth]{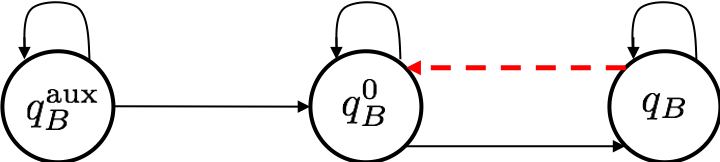}
    \caption{Graph $\ccalG$ constructed for the NBA shown in Figure \ref{fig:feas1}. The red dashed line corresponds to an accepting edge. Also, we have that $\ccalV_F=\{q_B\}$, $d_F(q_B^{\text{aux}},\ccalV_F)=2$, $d_F(q_B^{0},\ccalV_F)=1$, and $d_F(q_B,\ccalV_F)=0$.
    }
    \label{fig:graphG}
\end{figure}

Using the pruned NBA and the decomposable NBA transitions, we define the directed graph $\ccalG=\{\ccalV,\ccalE\}$ where $\ccalV\subseteq\ccalQ_B$ is the set of nodes and $\ccalE\subseteq\ccalV\times\ccalV$ is the set of edges. The set of nodes is defined so that $\ccalV=\ccalD_{q_B^{\text{aux}}}$ and the set of edges is defined so that $(q_B,q_B')\in\ccalE$ if there exists a feasible symbol that incurs the run $\rho_w$ defined in \eqref{eq:run}; see also Fig. \ref{fig:graphG}. 

\begin{figure}
    \centering
     \includegraphics[width=1\linewidth]{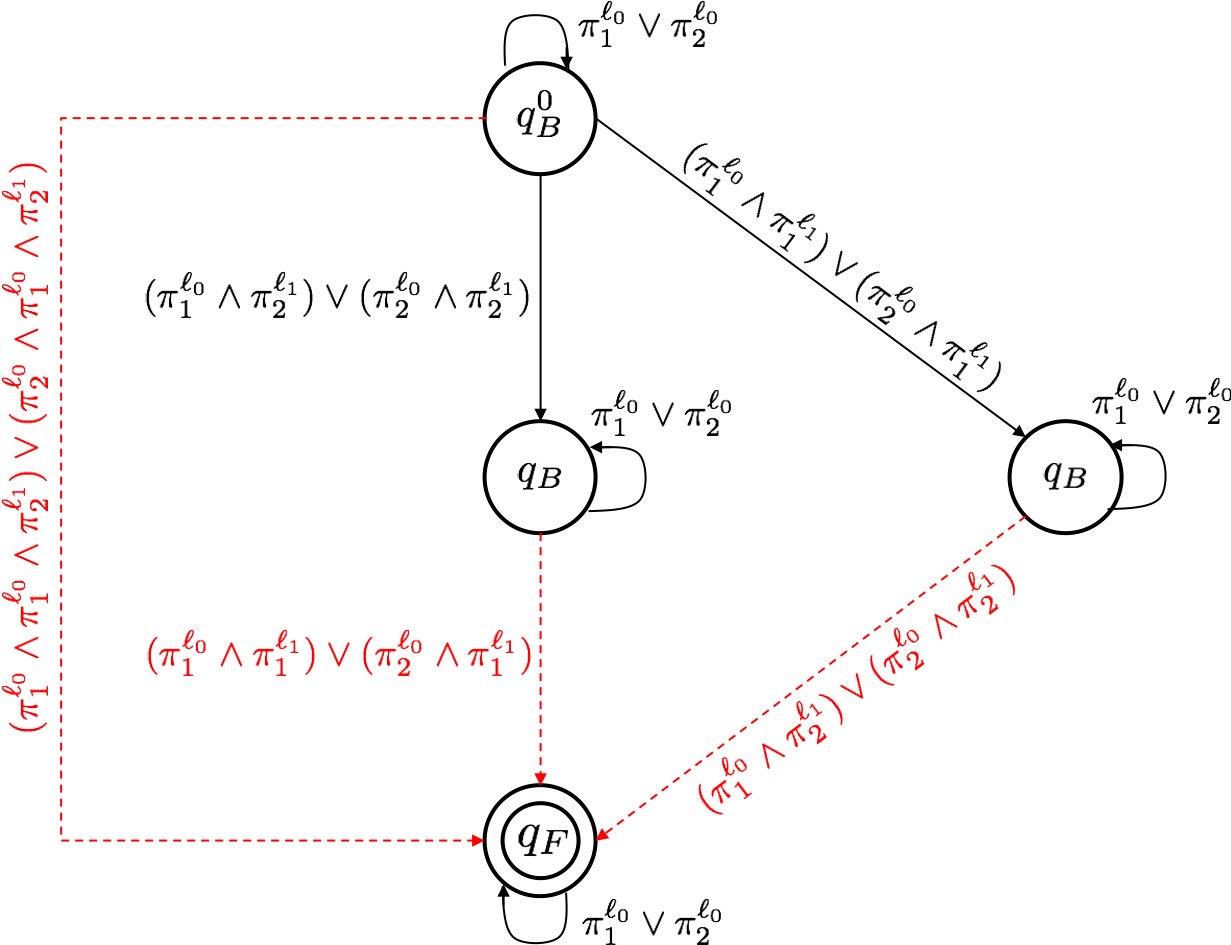}
    \caption{Decomposing global LTL tasks into local reachability tasks is not always possible. This figure shows the NBA corresponding to $\phi=\square(\pi_1^{\ell_0}\vee\pi_2^{\ell_0}) \wedge \Diamond \pi_2^{\ell_1}\wedge \Diamond \pi_2^{\ell_1}$. This formula requires either robot $1$ or $2$ to always be in $\ell_0$ and eventually robots $1$ and $2$ to visit, not necessarily simultaneously, $\ell_1$. Transition from $q_B^0$ to $q_B^F$ is removed as it requires robots to visit more than one location at the same time. Observe that the transitions from $q_B$ to $q_F$ and from $q_B'$ to $q_F$ do not satisfy  Definition \ref{def:feas}, i.e., $\ccalQ_{q_B}^{\text{next}}=\{q_B\}$ and $\ccalQ_{q_B'}^{\text{next}}=\{q_B'\}$, resulting in a disconnected graph $\ccalG$. For instance, consider the transition from $q_B$ to $q_F$. 
    Observe that transition from $q_B$ to $q_F$ cannot be cast as two independent reachability-avoidance problems. In fact, it requires coordination among the robots that is specific to the Boolean formula $b^{q_B,q_F}$. Specifically, when the robots reach $q_B$, we have that robot $1$ is in $\ell_0$ and robot $2$ is in $\ell_1$. To eventually satisfy $b^{q_B,q_F}=(\pi_1^{\ell_0}\wedge\pi_2^{\ell_1})\vee(\pi_2^{\ell_0}\wedge\pi_2^{\ell_1})$, first robot $2$ is required to return in $\ell_0$ while in the meantime robot $1$ should stay in $\ell_0$. Once robot $2$ returns in $\ell_0$, robot $1$ should move to $\ell_1$ enabling the transition to the final state $q_F$.}
    \label{fig:conserv}
\end{figure}

\begin{rem}(Definition \ref{def:feas})
Consider any transition $\delta_B^{m}(q_B,q_B')$ associated with the finite run \eqref{eq:run}. If $\delta_B^{m}(q_B,q_B')$ does not satisfy Definition \ref{def:feas} then this means that the transition from $q_B$ to $q_B'$ may be either feasible to enable but not in a local-reachability fashion (see Figure \ref{fig:conserv}) or impossible to activate as it may require robots to jump from one region of interest to another one (see Figure \ref{fig:NBA1}). 
\end{rem}


\section{Proofs of Lemmas}\label{sec:prop}

\subsection{Proof of Lemma \ref{lem:fragmentLTL}}
To show this result, first we express the Boolean formulas $b^{q_B,q_B'}$ in \eqref{eq:b} 
in CNF, i.e., as a conjunction of $F$ Boolean formulas $b_f$, for some $F>0$, where $b_f$ is a disjunction of literals, i.e.,
\begin{equation}\label{eq:bqbqb}
    b^{q_B,q_B'}=\bigwedge_{f=1}^F b_f.
\end{equation}
By Assumption \ref{as:cnf} we have that every clause $b_f$ is associated with only one robot, for all $f\in\{1,\dots,F\}$, i.e., $b_f$ can be written as a disjunction of atomic predicates $\pi_j^{r_e}$ that refer to the same robot $j$. Let $\ccalF_j$ be a set that collects all indices $f\in\{1,\dots,F\}$ to formulas $b_f$ that are associated with robot $j$, where $\ccalF_j\cap\ccalF_i=\emptyset$, for all $i,j\in\ccalN$, by definition of $b_f$. Next we define the Boolean formula 
\begin{equation}\label{eq:dj}
    d_j^{q_B,q_B'}=\bigwedge_{j\in\ccalF_j}b_f,
\end{equation}
which is a conjunction of all Boolean formulas $b_f$ associated with robot $j$ that appear in \eqref{eq:bqbqb}. Using \eqref{eq:dj} and the fact that $\ccalF_j\cap\ccalF_i=\emptyset$, for all $i,j\in\ccalN$, we can re-write \eqref{eq:bqbqb} as 
\begin{equation}\label{eq:bqbqb2}
    b^{q_B,q_B'}=\bigwedge_{j=1}^N d_j^{q_B,q_B'}.
\end{equation}
Thus, if Assumption \ref{as:cnf} holds, then the Boolean formula $b^{q_B,q_B'}$ defined in \eqref{eq:b} can be decomposed into local/decoupled sub-formulas $d_j^{q_B,q_B'}$ as in \eqref{eq:bqbqb2}. In other words, robot $j$ can make decisions to contribute to satisfaction of $b^{q_B,q_B'}$ independent from other robots by considering only satisfaction of the sub-formula $d_j^{q_B,q_B'}$. 

Due to the decomposition of \eqref{eq:b} as per \eqref{eq:bqbqb2}, checking if Definition \ref{def:feas} holds for a transition $\delta_B^m(q_B,q_B')$ is equivalent to checking if a `local' version of Definition \ref{def:feas} holds for all robots $j\in\ccalN^{q_B,q_B'}$ defined as follows. Specifically, a transition $\delta_B^m(q_B,q_B')$ is decomposable if the following condition holds for all robots $j\in\ccalN^{q_B,q_B'}$. For all symbols $\sigma^{q_B,q_B}_j\in\Sigma^{q_B,q_B}_j$, there should exist at least one symbol $\sigma^{q_B,q_B'}_j$ that satisfies $d_j^{q_B,q_B'}$ (i) and either (A) $j\notin\ccalR^{q_B,q_B}\cap\ccalR^{q_B,q_B'}$ or (B) $j\in\ccalR^{q_B,q_B}\cap\ccalR^{q_B,q_B'}$ and $L_{j}^{q_B,q_B}=L_{j}^{q_B,q_B'}$. 
Assume that there exists at least one robot $j$ for which this condition is not satisfied i.e., there exists at least one symbol $\sigma^{q_B,q_B}_j\in\Sigma^{q_B,q_B}_j$ for which there are no such $\sigma^{q_B,q_B'}_j$. This means that there exists $j\in\ccalR^{q_B,q_B}\cap\ccalR^{q_B,q_B'}$ and $L_{j}^{q_B,q_B'}\neq L_{j}^{q_B,q_B}$. Since $j\in\ccalR^{q_B,q_B}$, we have that transition towards the state $q_B$ was enabled because robot $j$ reached a region of interest $L_{j}^{q_B,q_B}$. Also, since $j\in\ccalR^{q_B,q_B'}$, we get that transition from $q_B$ to $q_B'$ will be enabled only if robot $j$ visits a region of interest denoted by $L_{j}^{q_B,q_B'}$. By assumption we have that $L_{j}^{q_B,q_B'}\neq L_{j}^{q_B,q_B}$. Therefore, we conclude that if Definition \ref{def:feas} is not satisfied then transition from $q_B$ to $q_B'$ can be enabled only if robot $j$ keeps being in a region of interest $L_{j}^{q_B,q_B}$ until it reaches another region $L_{j}^{q_B,q_B'}\neq L_{j}^{q_B,q_B}$, which is impossible given that all regions of interest are disjoint. In other words, under Assumption \ref{as:cnf}, if a multi-hop transition $\delta_B^m(q_B,q_B')$ does not satisfy Definition \ref{def:feas}, then this transition requires robots to jump from one region of interest to another which is impossible to achieve in practice completing the proof.

\subsection{Proof Of Lemma \ref{lem:binv}}

For simplicity of notation, let $b^{\text{inv}}$ and $b^{\text{next}}$ denote the Boolean formulas $b^{q_B,q_B}$ and $b^{q_B,q_B^{\text{next}}}$, respectively, i.e., the formulas that should be true to enable a transition from $q_B$ to $q_B$ and from $q_B$ to $q_B^{\text{next}}$, respectively. Once the robots reach $q_B(t)$, they coordinate to select a symbol $\sigma^{\text{next}}$ that satisfies $b^{\text{next}}$. By assumption, all robots will eventually reach their assigned locations $L_{j}^{\text{next}}$ but not necessarily at the same time instant. Once all robots reach their corresponding goal regions, at a time instant $T$, the symbol $\sigma^{\text{next}}$ that satisfies $b^{\text{next}}$ is generated.
Thus, to show that at time $T$ the transition from $q_B$ to $q_B^{\text{next}}$ will be enabled, it suffices to show that $b^{\text{inv}}$ is satisfied for all $t'\in [t,T)$ by construction of the NBA (or equivalently the graph $\ccalG$).\footnote{Note that the transition from $q_B$ to $q_B^{\text{next}}$ is, in general, a multi-hop NBA transition; see \eqref{eq:run}. Thus, formally, if the length of the this transition is $K$ hops, then once the robots generate $\sigma^{\text{next}}$ they should stay idle at their current locations $K-1$ time steps to reach $q_B^{\text{next}}$. Therefore, formally, transition to $q_B^{\text{next}}$ will occur at the time instant $T+K-1$.}



%
By construction of Algorithm \ref{alg:RRT}, we have that once the robots reach any state in $q_B\in\ccalV$, the corresponding Boolean formula $b^{\text{inv}}$ is satisfied by definition of \eqref{eq:b}. Thus, at time $t$ the multi-robot state $\bbp(t)$ (along with the map $M(t)$) satisfies the Boolean formula $b^{\text{inv}}$. As a result, if all robots in the set $\ccalN^{\text{inv}}$ (i.e., the robots that appear in $b^{\text{inv}}$) remain idle, the Boolean formula $b^{\text{inv}}$ will remain satisfied. Recall that robots $j\in\ccalN^{\text{next}}$ navigate the environment by following paths design as per \eqref{eq:localProb} 
while it may hold that $\ccalN^{\text{next}}\cap\ccalN^{\text{inv}}\neq \emptyset$. In what follows, we show that $q_B(t)=q_B(t')$ for all $t'\in[t,T)$. 
To show this, first recall that the robots $j\in\ccalR^{\text{inv}}\subseteq\ccalN^{\text{inv}}$ stay idle by construction of Alg. \ref{alg:RRT} and, therefore, they cannot violate $b^{\text{inv}}$. Thus, it suffices to show that the robots $j\in\bar{\ccalR}^{\text{inv}}\cap\ccalN^{\text{next}}$, where $\bar{\ccalR}^{\text{inv}}=\ccalN^{\text{inv}}\setminus\ccalR^{\text{inv}}$ will not violate $b^{\text{inv}}$. To show the latter we consider the following two cases. First, assume that $\bar{\ccalR}^{\text{inv}}\cap\ccalN^{\text{next}}=\emptyset$. Then this means that all robots in $\ccalN^{\text{inv}}$ will stay idle and, therefore, $b^{\text{inv}}$ will not be violated. Second, assume that there exist robots $j\in\bar{\ccalR}^{\text{inv}}\cap\ccalN^{\text{next}}$. By construction of the set $\bar{\ccalR}^{\text{inv}}$ we have that atomic predicates $\pi_j^{\ell_e}$ associated with robots $j\in\bar{\ccalR}^{\text{inv}}$ and for some regions $\ell_e$ appear in $b^{\text{inv}}$ only with a negation operator in front of them (i.e., in the form $\neg\pi_j^{\ell_e}$). In other words, robots $j\in\bar{\ccalR}^{\text{inv}}\cap\ccalN^{\text{next}}$ will violate $b^{\text{inv}}$ only if they visit certain regions of interest as they navigate the environment. 
Nevertheless, this is precluded by the first constraint in \eqref{eq:localProb} meaning that $q_B(t)=q_B(t')$ for all $t'\in[t,T)$. 
Finally, note that this result holds even if the robots execute their paths asynchronously, i.e., if the (continuous) time required to move from $\bbp_j(t)$ to $\bbp_j(t+1)$ is different across the robots. The reason is that (i) $\sigma_{\text{next}}$ will be generated eventually independent from the order where the robots arrive at their goal locations $L_{j}^{\text{next}}$ as long as all robots wait for each other once they arrive at $L_{j}^{\text{next}}$, (ii) satisfaction of $b^{\text{inv}}$ cannot be affected by the asynchronous execution of paths $\bbp_{j,t:T_j}$,  $j\in\ccalN^{\text{next}}$. The latter is true because, 
as discussed before, the (synchronous or asynchronous) actions of the robots $j\in\bar{\ccalR}^{\text{inv}}$ cannot violate $b^{\text{inv}}$ as long as they avoid certain regions, which is guaranteed by construction of the paths (see the first constraint in \eqref{eq:localProb}) completing the proof.

\subsection{Proof Of Proposition \ref{thm:completeness}}

To show this result it suffices to show that eventually the accepting condition of the NBA is satisfied, i.e., the robots will visit at least one of the final NBA states $q_F$ infinitely often. Equivalently, as discussed on Section \ref{sec:dist}, it suffices to show that accepting edges $(q_B,q_B')\in\ccalE$, where $q_B,q_B'\in\ccalV$ are traversed infinitely often. 

First, consider an infinite sequence of time instants $\bbt=t_0,t_1,\dots,t_k,\dots$ where $t_{k+1}\geq t_k$, so that an edge in $\ccalG$, defined in Section \ref{sec:dist}, is traversed at every time instant $t_k$. Let $e(t_k)\in\ccalE$ denote the edge that is traversed at time $t_k$. Thus, $\bbt$ yields the following sequence of edges $\bbe=e(t_0),e(t_1),\dots,e(t_k)\dots$ where $e(t_k)=(q_B(t_{k}),q_B(t_{k+1}))$, $q_B(t_0)=q_B^{\text{aux}}$, $q_B(t_k)\in\ccalV$, and the state $q_B^{k+1}$ is defined based on the following two cases. 
If $q_B(t_k)\notin \ccalV_F$, then the state $q_B(t_{k+1})$ is closer to $\ccalV_F$ than $q_B(t_k)$ is, i.e., $d_F(q_B(t_{k+1}),\ccalV_F)=d_F(q_B(t_{k}),\ccalV_F)-1$, where $d_F$ is defined in \eqref{eq:distF}. If $q_B(t_{k})\in\ccalV_F$, then $q_B(t_{k+1})$  is selected so that an accepting edge originating from $q_B(t_{k})$ is traversed. By definition of $q_B(t_{k})$, the `distance' to $\ccalV_F$ decreases as $t_k$ increases, i.e., given any time instant $t_k$, there exists a time instant $t_k'\geq t_k$ so that $q_B(t_{k}')\in\ccalV_F$ and then at the next time instant an accepting edge is traversed. This means that $\bbe$ includes an infinite number of accepting edges. 

In what follows, we show that that such a sequence exists in $\ccalG$ and Algorithm \ref{alg:RRT} will find it.
Assume that there exists a solution to Problem \ref{prob} and that the robots navigate the environment as per Algorithm \ref{alg:RRT} and update the graph $\ccalG$ by removing edges as described in Section \ref{sec:replan}. To show that Algorithm \ref{alg:RRT} is complete, it suffices to show that it can generate a infinite sequence of edges $\bbe$ as defined before.
%
We show the existence of such a sequence by induction. Specifically, we show that if there exists a time instant $t_k$ when the robots reach an NBA state denoted by $q_B(t_k)\in\ccalV$ there exists a time instant $t_{k+1}\geq t_k$ when the robots will reach a new state $q_B(t_{k+1})=q_B^{\text{next}}\in\ccalV$ where $q_B(t_{k+1})$ satisfies the conditions discussed before. 
At time $t=0$, we have that $q_B(0)=q_B^{\text{aux}}\in\ccalV$. By construction of Algorithm \ref{alg:RRT} the robots select a new state $q_B^{\text{next}}\in\ccalV$ that they should reach next in the same way as discussed before. By assumption (a), we have that there exists at least one NBA state $q_B^{\text{next}}$ that is reachable by $q_B(0)$. In other words, that there exists at least one symbol $\sigma^{\text{next}}$ associated with at least one decomposable transition $\delta_B^m(q_B(0),q_B^{\text{next}})$ so that the corresponding local planning problem \eqref{eq:localProb} is feasible for all robots $j$.
Then due to Lemma \ref{lem:binv}, we have that if the robots follow paths $\bbp_{j,t:T_j}$ that solve \eqref{eq:localProb}, as discussed in Section \ref{sec:exec}, the robots will eventually reach their respective regions $L_j^{\text{next}}$, i.e., the symbol $\sigma^{\text{next}}$ will be generated,  and the transition from $q_B(0)$ to $q_B^{\text{next}}$ will be enabled. 
%
%
Also, as the robots follow the paths as per Algorithm \ref{alg:RRT}, they will never hit an obstacle due to assumption (c). Particularly, by assumption (b) the robot is equipped with an omnidirectional and perfect sensor. 
Thus, if the path $\bbp_{j,t:T_j}$ passes through an obstacle, robot $j$ will always detect it and will re-solve \eqref{eq:localProb} to update its path. 
As a result, 
we get that there exists a time instant $t_0\geq 0$ when the robots will reach a new state $q_B^{\text{next}}\in\ccalV$. The induction step follows. Assume that at a time instant $t_{k}$ the robots reach a state $q_B(t_{k})\in\ccalV$ from a state $q_B(t_{k-1})\in\ccalV$. Following the same logic, we can show that there exists a time instant $t_{k+1}\geq t_k$ when the robots reach $q_B(t_{k+1})\in\ccalV$ that is closer to the final states than $q_B(t_{k})$ is, completing the proof.

\subsection{Proof of Proposition \ref{thm:exist}}
This result can be shown by following the same logic as in the proof of Proposition \ref{thm:completeness}. Specifically, by assumption (a) we have that there exist obstacle-free paths that respect the robot dynamics connecting reachable regions in the environment that can generate an infinite sequence of edges, defined as $\bbe=e_0,e_1,\dots,e_k\dots$, in $\ccalG$ that includes an infinite number of accepting edges meaning that a solution a solution to Problem \ref{prob} exists.


\bibliographystyle{IEEEtran}
\bibliography{YK_bib.bib}

\end{document}